\documentclass{article}

\usepackage{microtype}
\usepackage{graphicx}
\usepackage{subfigure}
\usepackage{booktabs} 

\usepackage{hyperref}

\usepackage[accepted]{icml2023}

\usepackage{amsmath}
\usepackage{amssymb}
\usepackage{mathtools}
\usepackage{amsthm}

\usepackage[capitalize,noabbrev]{cleveref}

\theoremstyle{plain}
\newtheorem{theorem}{Theorem}[section]

\theoremstyle{definition}
\newtheorem{definition}[theorem]{Definition}
\newtheorem{assumption}[theorem]{Assumption}
\theoremstyle{remark}

\usepackage[textsize=tiny]{todonotes}

\icmltitlerunning{Adversarial Example Does Good: Preventing Painting Imitation from Diffusion Models via Adversarial Examples}

\begin{document}

\twocolumn[
\icmltitle{Adversarial Example Does Good: Preventing Painting Imitation from \\Diffusion Models via Adversarial Examples}



\icmlsetsymbol{equal}{*}

\begin{icmlauthorlist}
\icmlauthor{Chumeng Liang}{equal,1,2}
\icmlauthor{Xiaoyu Wu}{equal,1,2}
\icmlauthor{Yang Hua}{3}
\icmlauthor{Jiaru Zhang}{1}
\icmlauthor{Yiming Xue}{4,2}
\icmlauthor{Tao Song}{1}
\icmlauthor{Zhengui Xue}{1}
\icmlauthor{Ruhui Ma}{1}
\icmlauthor{Haibing Guan}{1}
\end{icmlauthorlist}
\icmlaffiliation{1}{Shanghai Jiao Tong University, China}
\icmlaffiliation{2}{cheer4creativity.ai}
\icmlaffiliation{3}{Queen’s University Belfast, UK}
\icmlaffiliation{4}{NYU, USA}

\icmlcorrespondingauthor{Tao Song}{songt333@sjtu.edu.cn}
\icmlkeywords{Machine Learning, ICML}

\vskip 0.3in
]



\printAffiliationsAndNotice{\icmlEqualContribution} 


\begin{abstract}
Recently, Diffusion Models (DMs) boost a wave in AI for Art yet raise new copyright concerns, where infringers benefit from using unauthorized paintings to train DMs to generate novel paintings in a similar style. 
To address these emerging copyright violations, in this paper, we are the first to explore and propose to utilize adversarial examples for DMs to protect human-created artworks. Specifically, we first build a theoretical framework to define and evaluate the adversarial examples for DMs. Then, based on this framework, we design a novel algorithm, named AdvDM, which exploits a Monte-Carlo estimation of adversarial examples for DMs by optimizing upon different latent variables sampled from the reverse process of DMs. 
Extensive experiments show that the generated adversarial examples can effectively hinder DMs from extracting their features. Therefore, our method can be a powerful tool for human artists to protect their copyright against infringers equipped with DM-based AI-for-Art applications. The code of our method is available on GitHub:~\url{https://github.com/mist-project/mist.git}.
\end{abstract}

\section{Introduction}
\label{sec:intro}
Recent years have witnessed a boom of deep diffusion models~\cite{sohl2015deep,ho2020denoising} in computer vision. With solid theoretical foundations~\cite{song2020score,song2020denoising,bao2022analytic} and highly applicable techniques~\cite{gal2022image,lu2022dpm}, diffusion models have proven to be effective in generative tasks, including image synthesis~\cite{ruiz2022dreambooth}, video synthesis~\cite{yang2022diffusion}, image editing~\cite{kawar2022imagic}, and text-to-3D synthesis~\cite{poole2022dreamfusion}. Among these, the latent diffusion model~\cite{rombach2022high} shows great power in artwork creation, sparking a commercialization flurry of AI for Art.

\begin{figure}[t]
\vspace{-0.2cm}
\begin{center}
\includegraphics[width=\columnwidth]{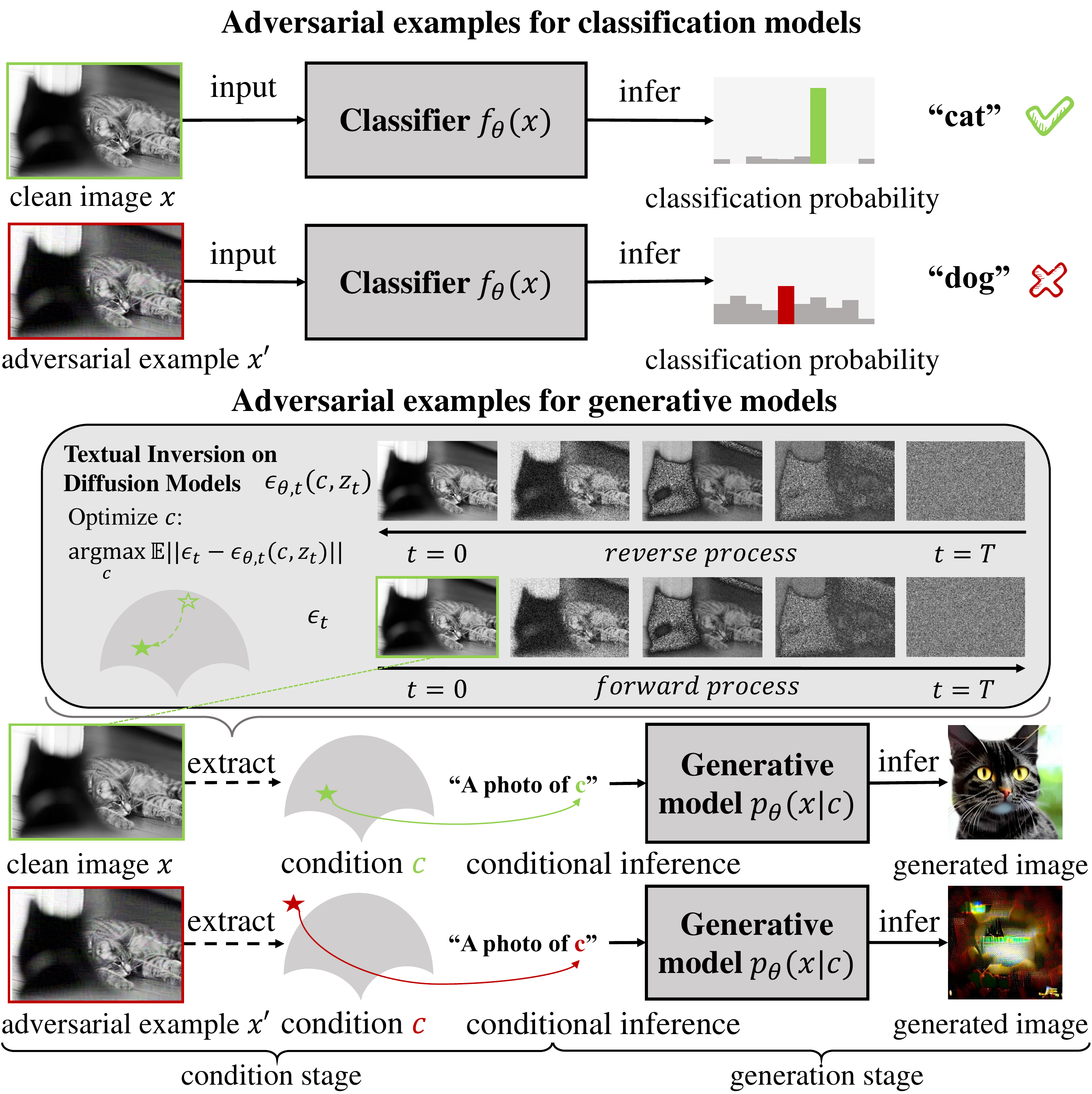}
\caption{Comparison of workflows for adversarial examples in classification models and diffusion models. Adversarial examples in diffusion models prevent diffusion models from extracting image features as conditions by inducing out-of-distribution features. The feature extracting shown in the figure is textual inversion~\cite{gal2022image} in DMs, which has raised copyright concerns in several cases~\cite{kimjungginews,mimicnews}. }
\label{fig:framework}
\end{center}
\vspace{-0.4cm}
\end{figure}

Despite the success of diffusion models in commercialization, it has been a public concern that these models empower some copyright violations. For example, \textit{textual inversion}~\cite{gal2022image}, a novel function implemented in most AI-for-Art applications based on the latent diffusion model, can imitate the art style of human-created paintings with several samples. Cases have taken place that copyright infringers easily fetch paintings created by artists online and illegally use them to train models with the help of textual inversion~\cite{kimjungginews,mimicnews}. Although artists have the right to declare prohibition for their artworks to be used for training AI-for-Art models, there is no existing technology to prevent or track this illegal use, leading to an even lower crime cost and difficulty in proof generating. Moreover, artists suffer from a lack of resources to start legal challenges against the infringers~\cite{legalnews} (See Appendix~\ref{sec:ethicissues} for more discussion about ethical issues of AI for Art). Hence, the art society is calling for off-the-shelf techniques to protect the copyright of paintings against AI for Art~\cite{legalnews}.

Inspired by the adversarial examples in image classification~\cite{fgsm,pgd,cw}, an idea for this protection is to add some tailored and tiny perturbations to images and make them \textit{unrecognizable} for the diffusion model in AI-for-Art applications. Here, \textit{unrecognizable} means the image cannot be recognized as a normal image by the diffusion model and hence restrains the model from extracting image features or imitating the art style. We consider these perturbed images as \textit{adversarial examples for diffusion models}. By transferring paintings into adversarial examples without losing the image semantics, the lack of techniques in artwork copyright protection can be resolved.

However, generating adversarial examples for diffusion models is non-trivial. Unlike classification models, diffusion models exploit input images by generating new images conditioned on the inputs rather than conducting an end-to-end inference on them. An adversarial example must then prevent its feature (e.g., styles, contents, \textit{etc}.) from being extracted in some identifiable conditions by the diffusion model. Furthermore, the training objective of diffusion models is optimized indirectly through a variational bound and thus is not applicable in the optimization of the adversarial example. For these reasons, existing research only exploits diffusion models to improve the robustness of classifiers~\cite{nie2022diffusion}, leaving a blank in the formulation of adversarial examples for diffusion models. 

In this paper, we build a theoretical framework to define and evaluate the adversarial example for diffusion models. Specifically, adversarial examples work on protecting their own feature from being extracted in the inference workflow of diffusion models (See in Figure~\ref{fig:framework}). This workflow consists of two stages: (1) the condition stage that extracts the feature from input images as conditions, and (2) the generation stage that generates images based on these conditions. In the case shown in Figure~\ref{fig:framework}, the condition stage is empowered by textual inversion~\cite{gal2022image}. Our adversarial examples work by misleading the feature extracting in the condition stage and resulting in an out-of-distribution condition. To this end, we define the adversarial example with an optimization target to minimize the probability that the image is recognized as a real image by the diffusion model. We optimize the target by adding tiny perturbations to the image. We then formulate the evaluation for the adversarial example according to the workflow shown in Figure~\ref{fig:framework}. A good adversarial example would result in a bad quality of conditional generated images, by which we can evaluate the quality of adversarial examples.

Under the proposed framework, we propose an algorithm to generate the adversarial example for diffusion models. We conduct a Monte Carlo method to estimate the objective function given by our definition in the context of diffusion models. We also evaluate our adversarial examples with real copyright violation scenarios~\cite{kimjungginews,mimicnews}. Extensive experiments show that our adversarial examples can efficiently hinder the latent diffusion model used by commercialization applications from extracting their features and imitating their styles or contents.

Our contributions are summarized in the following aspects.
\begin{itemize}
\vspace{-0.2cm}
    \item We construct a novel framework to define and evaluate the adversarial examples for diffusion models. To the best of our knowledge, we are the first to systematically investigate this topic.
    \item Under the above framework, we propose an end-to-end algorithm AdvDM to generate the adversarial examples for diffusion models.
    \item We conduct extensive experiments on several datasets, covering single-category and art-style ones, to validate that our method can effectively protect images from being learned, imitated, and copied by diffusion models.
\vspace{-0.2cm}
\end{itemize}

\section{Background}
\label{sec:bg}
\subsection{Generative Modeling and Diffusion Models}
A generative model learns from data $x\sim q(x)$ and holds a distribution $p_{\theta}(x)$ where generated data can be sampled. Generative models based on latent variables have proven effective in generative tasks, including VAEs \cite{DBLP:journals/corr/KingmaW13,razavi2019generating} and GANs~\cite{goodfellow2014generative,brock2018large}. These models match data with a \textit{latent variable} $z$ in low-dimensional space and model the joint distribution $p_{\theta}(x,z)$.

An intuitive idea to train a generative model is to maximize $p_{\theta}(x)$ for real data $x\sim q(x)$. However, $p_{\theta}(x)$ is difficult to optimize directly thus requiring transformation, where the variational bound~\cite{higgins2016beta,gregor2016towards} given by $-\log p_{\theta}(x)\leq -\log\frac{p_{\theta}(x,z)}{q(z|x)}$ is selected to be optimized instead.

An important paradigm in generative models is conditional generative modeling~\cite{mirza2014conditional}. Generally, conditional generative models use different forms of conditions to do image generation, including categories~\cite{mirza2014conditional}, base images~\cite{zhu2017unpaired}, characteristics~\cite{karras2019style,karras2020analyzing}, and condition prompting in natural languages~\cite{rombach2022high}. Denoting the condition by $c$, they model $p_{\theta}(x|c)$ with parameter $\theta$ and support sampling images $x$ from this distribution.

As the generative model defining the state-of-the-art, diffusion models~\cite{sohl2015deep,ho2020denoising} construct a series of latent variables $x_{1:T}$ by a Markov Chain $q(x_{1:T}|x_{0})$. A reverse Markov Chain $p_{\theta}(x_{0:T})$ is then used to revert the latent variables to the data $x_0$. $p_{\theta}(x)$ is optimized with the variational bound of $p_{\theta}$, 
\begin{equation}
\label{eq:dmloss}
\begin{aligned}
    -\log p_{\theta}(x)\leq -\log\frac{p_{\theta}(x_{0:T})}{q(x_{1:T}|x_0)} := \mathcal{L}_{DM}.
\end{aligned}
\end{equation}
Intuitively, DMs generate images by learning to recover an image from noise by the denoising reverse process $p_{\theta}(x_{0:T})$. A recent breakthrough has taken place when researchers deploy the denoising reverse process in a latent space~\cite{rombach2022high}. This \textit{latent diffusion model} (LDM) has achieved state-of-the-art performance in both image quality in artwork generation and sampling efficiency, being the mainstream model used in AI-for-Art applications. 


\subsection{Adversarial Examples}

Let $p_{data}(y,x)$ denote the joint distribution between data $x$ and label $y$. A classification model with parameter $\theta$ is  expected to estimate $p_{data}(y|x)$ with $p_{\theta}(y|x)$. The difference between the two distributions can be quantitated by KL Divergence. The optimization goal of the classification model can be then formulated by 

\begin{equation}
\label{eq:classifier}
\begin{aligned}
    \arg\min\limits_{\theta} KL(p_{data}(y|x)||p_{\theta}(y|x)).
\end{aligned}
\end{equation}

Various terms of loss function are exploited as alternative optimization targets to this goal in classification modeling. The maximum log-likelihood, a widely-used loss function, proves to be equivalent to the goal in Eq.~(\ref{eq:classifier})~\cite{shlens2014notes}. To simplify the notation, we use $\mathcal{L}_{\theta}(x, y)$ to denote the loss term and minimize it in the optimization. 

The neural network used in the classification model is vulnerable to \textit{adversarial examples}: given an input $x$ and its label $y$, it is possible to find a new input $x'$ not classified to $y$~\cite{fgsm,cw}. The adversarial example $x'$ for classification models can be formulated by 

\begin{equation}
\label{eq:adv_classifier}
\begin{aligned}
    x':=\arg\max\limits_{x'} \quad &\mathcal{L}_{\theta}(x'),\\
    s.t.\quad||x-x'&||\leq \epsilon.
\end{aligned}
\end{equation}


Various research investigates adversarial examples related to generative models. They mainly consider adversarial examples generated by generative models and misleading classification models~\cite{kos2018adversarial}. A discussion of adversarial attacks on flow-based generative models ~\cite{pope2020adversarial} aims at finding examples similar to real images but with low likelihood scores in flow-based models. However, its theoretical analysis only considers cases in flow-based models and applies a strong assumption that the data $x$ is normally distributed. Moreover, it does not yield a general formulation of adversarial examples for generative models.

\section{Adversarial Examples for Diffusion Models}
\label{sec:method}
In this section, we discuss how to generate and evaluate adversarial examples for Diffusion Models (DMs). We first formulate the objective function, which minimizes the probability that the example is a real image and sampled by the model. Then, we propose AdvDM, an algorithm to approximately generate adversarial examples for DMs. Finally, we discuss the concrete evaluation for these adversarial examples. Following the notation of DMs~\cite{ho2020denoising}, the image $x$ is denoted by $x_0$ and the latent variable $z$ is denoted by $x_{1:T}$ in this section.

\subsection{Adversarial Examples for Diffusion Models}
\label{sec:3.1}
We consider adversarial examples that cannot be recognized as real images by diffusion models but are visibly similar to real images. For a diffusion model $\theta$, an adversarial example is out-of-distribution for the generated distribution $p_{\theta}(x)$. To generate such adversarial examples, an idea is to minimize $p_{\theta}(x+\delta)$ by adding one or several perturbations $\delta$ whose scale is strictly constrained. The constraint of the perturbation scale ensures the perturbation is human-invisible and does not hurt the image semantics. Based on this idea, we define the adversarial example for a diffusion model parameterized by $\theta$.
\begin{definition}[\textbf{Adversarial Example for Diffusion Models}]
\label{def:1}
Given a diffusion model parameterized by $\theta$ and the distribution of real data $q(x)$, the adversarial example $x'$ is formulated by $x'=x+\delta$, where $x\sim q(x)$ and $\delta$ is given by the following equation:
\end{definition}
\begin{equation}
\label{eq:definition}
\begin{aligned}
    \delta &:= \arg\min\limits_{\delta} p_{\theta}(x+\delta),\\
    &\mbox{where } x\sim q(x),\Vert\delta\Vert\leq\epsilon,\\
    \epsilon\mbox{ is}&\mbox{ a constant and usually small}.
\end{aligned}
\end{equation}

 However, $p_{\theta}$ is not practically computable in diffusion models. With the help of the latent variable $p_{\theta}(x)$, we can estimate $p_{\theta}(x+\delta)$ by Monte Carlo. To this end, we expand $p_{\theta}(x)$ over the latent variable $x_{1:T}$,
\begin{equation}
\label{eq:dmadv}
\begin{aligned}
     p_{\theta}(x)&=\int p_{\theta}(x_{0:T})\text{d}x_{1:T}.\\
\end{aligned}
\end{equation}
We denote the adversarial example $x+\delta$ by $x'$, with $\Vert\delta\Vert\leq\epsilon$. Eq.~(\ref{eq:dmadv}) suggests it is possible to minimize $p_{\theta}(x)$ by Monte Carlo: By minimizing $p_{\theta}(x_{0:T})$ with different sampling processes of $x_{1:T}$, we are approximately minimizing $p_{\theta}(x')$. Let $u(x'_{1:T})$ denote the distribution of $x'_{1:T}$. We can alter our optimization goal to the following form:

\begin{equation}
\label{eq:realtarget}
\begin{aligned}
    \delta &:= \arg\min\limits_{\delta} \mathbb{E}_{x'_{1:T}\sim u(x'_{1:T})}p_{\theta}(x'_{0:T}),\\
    &\mbox{where } x\sim q(x), x'=x+\delta.
\end{aligned}
\end{equation}
An advantage in DMs is that the posterior $q(x'_{1:T}|x'_0)$ is a Gaussian distribution with fixed parameters exactly independent from $x'_0$. Therefore, it is possible to regularize $p_{\theta}(x'_{0:T})$ with $q(x'_{1:T}|x'_0)$. We use the negative log term of $p_{\theta}(x'_{0:T})$ as usual. The final form of our objective function can be then inferred.
\begin{equation}
\label{eq:realtarget2}
\begin{aligned}
    &\min\limits_{\delta}\mathbb{E}_{x'_{1:T}\sim u(x'_{1:T})} p_{\theta}(x'_{0:T}) \\
    =& \max\limits_{\delta}\mathbb{E}_{x'_{1:T}\sim u(x_{1:T})} -\log p_{\theta}(x'_{0:T})\\
    =& \max\limits_{\delta}\mathbb{E}_{x'_{1:T}\sim u(x'_{1:T})} -\log\frac{p_{\theta}(x'_{0:T})}{q(x'_{1:T}|x'_0)}\\=& \max\limits_{\delta}\mathbb{E}_{x'_{1:T}\sim u(x'_{1:T})} \mathcal{L}_{DM}(x',\theta).
\end{aligned}
\end{equation}
Intuitively, Eq.~(\ref{eq:realtarget2}) generates adversarial example $x'$ by maximizing the loss used for training DMs with different latent variables sampled from $u(x'_{1:T})$.

\subsection{AdvDM: Generating Adversarial Examples by Monte Carlo}
\label{sec:3.2}
In this subsection, we propose AdvDM, the algorithm to generate adversarial examples for DMs. Inspired by existing methods of adversarial attack on classification tasks~\cite{fgsm,pgd,cw}, we exploit the gradient of our optimization goal. A difference is that we cannot analytically compute the gradient of the objective function $\mathbb{E}_{x_{1:T}\sim u(x_{1:T})} \mathcal{L}_{DM}(\theta)$ since it is the gradient of an expectation. As mentioned in Section~\ref{sec:3.1}, we estimate it by the expected gradient with Monte Carlo. For each iteration, we sample a $x'_{1:T}\sim u(x'_{1:T})$ and compute a gradient of $\mathcal{L}_{DM}(\theta)$ accordingly. We then do one step of gradient ascent with this gradient. The estimation is summarized in Eq.~(\ref{eq:estimatedgradient}):

\begin{equation}
\label{eq:estimatedgradient}
\begin{aligned}
    \nabla_{x_0}\mathbb{E}_{x_{1:T}\sim u(x_{1:T})} \mathcal{L}_{DM}(\theta)\approx \mathbb{E}_{x_{1:T}\sim u(x_{1:T})}\nabla_{x_0}\mathcal{L}_{DM}(\theta).
\end{aligned}
\end{equation}

We follow existing methods of adversarial attack ~\cite{fgsm,pgd} and apply a sign function to constrain the scale of the estimated gradient. Let $x_0^{(i)}$ denote the adversarial example of the $i$th step in optimization. The adversarial example of the $(i+1)$th step is generated by a signed gradient ascent with step length $\alpha$,

\begin{equation}
\label{eq:closeform}
    x_0^{(i+1)} = x_0^{(i)} +\alpha\mbox{sgn}(\nabla_{x_0^{(i)}}\mathcal{L}_{DM}(\theta)|_{x^{(i)}_{1:T}\sim u(x^{(i)}_{1:T})}), 
\end{equation}
where $\mbox{sgn}$ refers to the sign function.

Intuitively, AdvDM samples different latent variables and iteratively conducts one step of gradient ascent on the loss of DMs with different for each sampling. In practice, we let $u(x_{1:T})$ be the posterior $q(x_{1:T}|x_0)$, for it induces a good performance empirically in the experiment. We summarize AdvDM in Algorithm~\ref{alg:pdfai}.  The implementation details are shown in Appendix~\ref{details_advdm}.

\begin{algorithm}[htbp]
   \caption{AdvDM: Adversarial Example for DMs}
   \label{alg:pdfai}
\begin{algorithmic}
   \STATE {\bfseries Input:} Data $x_0$, parameter $\theta$, number of Monte Carlo $N$, step length $\alpha$
   \STATE {\bfseries Output:} Adversarial example $x'_0$
   \STATE Initialize $x_0^{(0)} \leftarrow x_0$.
   \FOR{$i=1$ {\bfseries to} $N$}
   \STATE Sample $x^{(i)}_{1:T}\sim q(x^{(i)}_{1:T}|x_0^{(i)})$
   \STATE $\delta^{(i)}\leftarrow\alpha\mbox{sgn}(\nabla_{x_0^{(i)}}\mathcal{L}_{DM}(\theta)|_{x^{(i)}_{1:T}})$
   \STATE $x_0^{(i)}\leftarrow x_0^{(i-1)}+\delta^{(i)}$
   \ENDFOR
   \STATE $x'_0\leftarrow x_0^{(N)}$
\end{algorithmic}
\end{algorithm}

\subsection{Evaluating the Quality of Adversarial Examples}
\label{sec:3.3}
The diffusion model $\theta$ is evaluated by the quality of images sampled from $p_{\theta}(x)$~\cite{goodfellow2014generative,ho2020denoising}. This sampling is called the \textit{inference} of the diffusion model. Unlike classification models, diffusion models do not take images as input directly but exploit them by extracting features from them and generating images conditioned on these features. We mainly focus our evaluation scenario on this conditional inference, where copyright violations have taken place. For unconditional inference, the model samples a noise and generates images. This process has no input images and does not raise copyright concerns, thus not included in our evaluation.

Following the existing research in adversarial examples~\cite{fgsm,graph,nlp}, we evaluate the adversarial example for diffusion models in inference by measuring how much it would hurt the performance of image generation. As shown in Figure~\ref{fig:framework}, the inference is divided into two stages. In the condition stage, the diffusion model extracts features from the input image. In the generation stage, the model exploits these features as conditions to generate new images. We denote the condition by $c$ and the feature-extracting process by $p_{\theta}(c|x)$. In practice, $c$ can be a prompting in natural language~\cite{gal2022image,ruiz2022dreambooth} that abstracts the image semantics or a latent variable~\cite{rombach2022high} related to the image.

The diffusion model $\theta$ can then generate an image $x_g$ with a condition $c_{g}$ sampled from $p_{\theta}(c|x)$. We model this process by $p_{\theta}(x|c_g)$, with $x_g\sim p_{\theta}(x|c_g)$. Note that $p_{\theta}(x|c_g) =p_{\theta}(x)\frac{p_{\theta}(c_g|x)}{p_{\theta}(c_g)}$. We assume a dependency between $c_g$ and image sample $x$.

\begin{assumption}[\textbf{Dependency between $c_g$ and $x$}]
\label{amp:2}
$c_g$ is a condition sampled from $p_{\theta}(c|x)$. We have $\frac{p_{\theta}(c_g|x)}{p_{\theta}(c_g)}\geq 1$.
\end{assumption}
Assumption~\ref{amp:2} is plausible for most cases since $\theta$ is a trained diffusion model and promises a strong relationship between samples and conditions semantically. 
With this assumption, $p_{\theta}(x|c)$ can be a higher bound of $p_{\theta}(x)$ and an alternative distribution for sampling. As the normal evaluation of diffusion models, we also evaluate $p_{\theta}(x|c)$ by applying a quality metric $\mathbf{D}(\cdot)$ to the image sampled from $p_{\theta}(x|c)$. The evaluation is summarized in Algorithm~\ref{alg:eva}.

\begin{algorithm}[htbp]
   \caption{Evaluating Adversarial Example for diffusion models}
   \label{alg:eva}
\begin{algorithmic}
   \STATE {\bfseries Input:} Adversarial example(s) $x_{adv}$, diffusion model $\theta$, sample quality metric $\mathbf{D}(\cdot)$
   \STATE {\bfseries Output:} the sample quality $\mathcal{Q}$
   \STATE Initialize the dataset $x_r \leftarrow x_{adv}$
   \STATE Sample $c_g\sim p_{\theta}(c|x_r)$
   \STATE Generate images by sampling $x_g\sim p_{\theta}(x|c_g)$
   \STATE $\mathcal{Q}\leftarrow\mathbf{D}(x_g, x_r)$ 
\end{algorithmic}
\end{algorithm}

A good adversarial example prevents $p_{\theta}(c|x)$ from extracting $c$ accurately and results in a bad sample quality of $x_g$, which can be measured by the sample quality metric $\mathbf{D}(\cdot)$. In practice, we select Fréchet Inception Distance (FID)~\cite{heusel2017gans} and Precision ($prec.$)~\cite{kynkaanniemi2019improved} as $\mathbf{D}(\cdot)$.   

Three scenarios of $p_{\theta}(c|x)$ and $p_{\theta}(x|c)$ are considered in the evaluation of adversarial examples for diffusion models as follows. They either have been~\cite{kimjungginews,mimicnews} or can be the scenario of copyright violations with AI for Art. Details of three scenarios are given in Appendix~\ref{scenario}.

\begin{enumerate}
    \item \textbf{Text-to-image generation based on textual inversion}: Given a small batch of images $x$ depicting objects of the same category, $p_{\theta}(c|x)$ abstracts the object in images with a word $S^{*}$ in natural language. Let the condition $c_g$ be $S^{*}$. This is often implemented by the language model embedded in the diffusion model. $p_{\theta}(x|c_g)$ then generates images $x_g$ conditioned on $S^{*}$ with the diffusion model. This scenario is shown in Figure~\ref{fig:framework}.
    \item \textbf{Style transfer based on textual inversion}: 
    Given a small batch of images $x$ depicting objects of the same art style, $p_{\theta}(c|x)$ abstracts the common art style of images with a word $S^{*}$ in natural language. Let $c_g$ be $S^{*}$. $p_{\theta}(x|c_g)$ then generates images $x_g$ conditioned on $S^{*}$ with the diffusion model. In practice, we start generation from step $t$ based on the latent variable $z_{s,t}$ at step $t$ from another source image $x_{s}$ for better visualization. The generation can be exactly formulated by $p_{\theta}(x|c_g,z_{s,t})$.
    \item \textbf{Image-to-image synthesis}: Given an image $x$, $p_{\theta}(c|x)$ samples a latent variable $z_t$ at the denoising step $t$. Let $c_g$ be $z_t$. We start the generation $p_{\theta}(x|c_g)$ from step $t$ based on $z_{t}$. 
\end{enumerate}

\begin{table*}[th]
\caption{Text-to-image generation based on textual inversion}
\label{tab:main-res}
\vskip 0.15in
\begin{center}
\begin{small}
\begin{sc}
\begin{tabular}{lccccccccc}
\toprule
Dataset & \multicolumn{3}{c}{LSUN-Cat} & \multicolumn{3}{c}{LSUN-Sheep} &  \multicolumn{3}{c}{LSUN-Airplane}\\
 Metric & FID $\uparrow$ & $prec.\downarrow$ & $recall.$ & FID $\uparrow$ & $prec.\downarrow$ & $recall.$ & FID $\uparrow$ & $prec.\downarrow$ & $recall.$\\
\midrule
No attack  &  34.94 & 0.5643 & 0.1531  & 32.81  & 0.6378 & 0.1228 &  39.22 & 0.5016& 0.2765\\
AdvDM   &  127.04 & 0.1708& 0.061 & 203.5& 0.0058 & 0.378 &  169.67& 0.0263& 0.3235\\
\bottomrule
\end{tabular}
\end{sc}
\end{small}
\end{center}
\vskip -0.1in
\end{table*}


\section{Experiment}
\label{sec:exp}

In this section, we evaluate our proposed AdvDM to generate adversarial examples for DMs. Since our motivation is to help protect paintings against being illegally used by AI-for-Art applications, we choose the Latent Diffusion Model~\cite{rombach2022high} (LDM)  backbone~\footnote{https://ommer-lab.com/files/latent-diffusion/nitro/txt2img-f8-large/model.ckpt},  which is the mainstream model used in AI-for-Art applications (see  Appendix~\ref{sec:stable-diffusion}). The implementation details for AdvDM on Latent Diffusion Model are shown in Appendix~\ref{details_advdm}. We fix $l_{\infty}$ norm as the constraint for generating all the adversarial examples. Following existing research in adversarial examples, we set the sampling step as 40, the per-step perturbation budget as 1/255, the total budget as 8/255, and the batch size as 4. We conduct experiments on categories of LSUN~\cite{yu2015lsun} and WikiArt~\cite{wikiart}. We use 8 NVIDIA RTX A4000 GPUs for all experiments. Visualization of all experiments is shown in Appendix~\ref{visualization}. Additionally, we evaluate AdvDM on more conditional generation tools based on the Latent Diffusion Model and demonstrate the results in Appendix~\ref{appendix:additional}. We also discuss other potential methods to generate adversarial examples for DMs in comparison with AdvDM. The results are listed in Appendix~\ref{attacks}.
\subsection{Text-to-image generation based on textual inversion}
\label{sam}
We first evaluate our adversarial examples on the text-to-image generation with textual inversion, as mentioned in Section \ref{sec:3.3}. To evaluate AdvDM quantitatively, we randomly select 1,000 images from LSUN-cat, LSUN-sheep, and LSUN-airplane. For all experimental settings, we follow the paper of textual inversion~\cite{gal2022image}. The images are separated into 5-image groups and we optimize a condition prompting $c$, i.e., pseudo-word $S^{*}$ in~\cite{gal2022image} for each group. $S^{*}$ is a word vector in the semantic space of the language model embedded in LDM~\cite{radford2021learning}, expected to capture the object in 5 images, e.g., cat for images in LSUN-cat. For each 5-image group and $S^{*}$, we set the iteration steps in optimization as 5,000 as default. We then use each pseudo word $S^{*}$ to generate 50 images conditionally, leveraging the text-to-image function of LDM, which results in a total of 10,000 generated images for each dataset. All the images are resized to $256\times  256$ as default. This generation process is conducted both on clean images and adversarial examples generated by AdvDM.

We evaluate the sample quality of generated images by two metrics: Fréchet Inception Distance (FID) and Precision ($prec.$), which both measure the similarity between generated images and training images. For images generated based on adversarial examples, a high FID and a low Precision show that these images cannot capture the object in the adversarial examples. As a reference, Recall ($recall$) is also calculated in our experiment. However, the difference between Recall of images generated based on clean images and of those on adversarial examples is unpredictable. This is because Recall measures the diversity of generated images rather than the similarity between generated and training images, which is out of the concern raised in our motivation. Implementation details of the metrics mentioned above are shown in Appendix~\ref{details_eva}.

The results are shown in Table~\ref{tab:main-res}. Our adversarial examples significantly increase FID and decrease Precision of the conditionally-generated images. Meanwhile, Recall does not vary consistently. This suggests our adversarial examples are powerful in protecting its contents from being extracted as generation conditions.

\begin{figure*}[ht]
\vspace{-0.1cm}
\begin{center}
\includegraphics[width=0.99\textwidth]{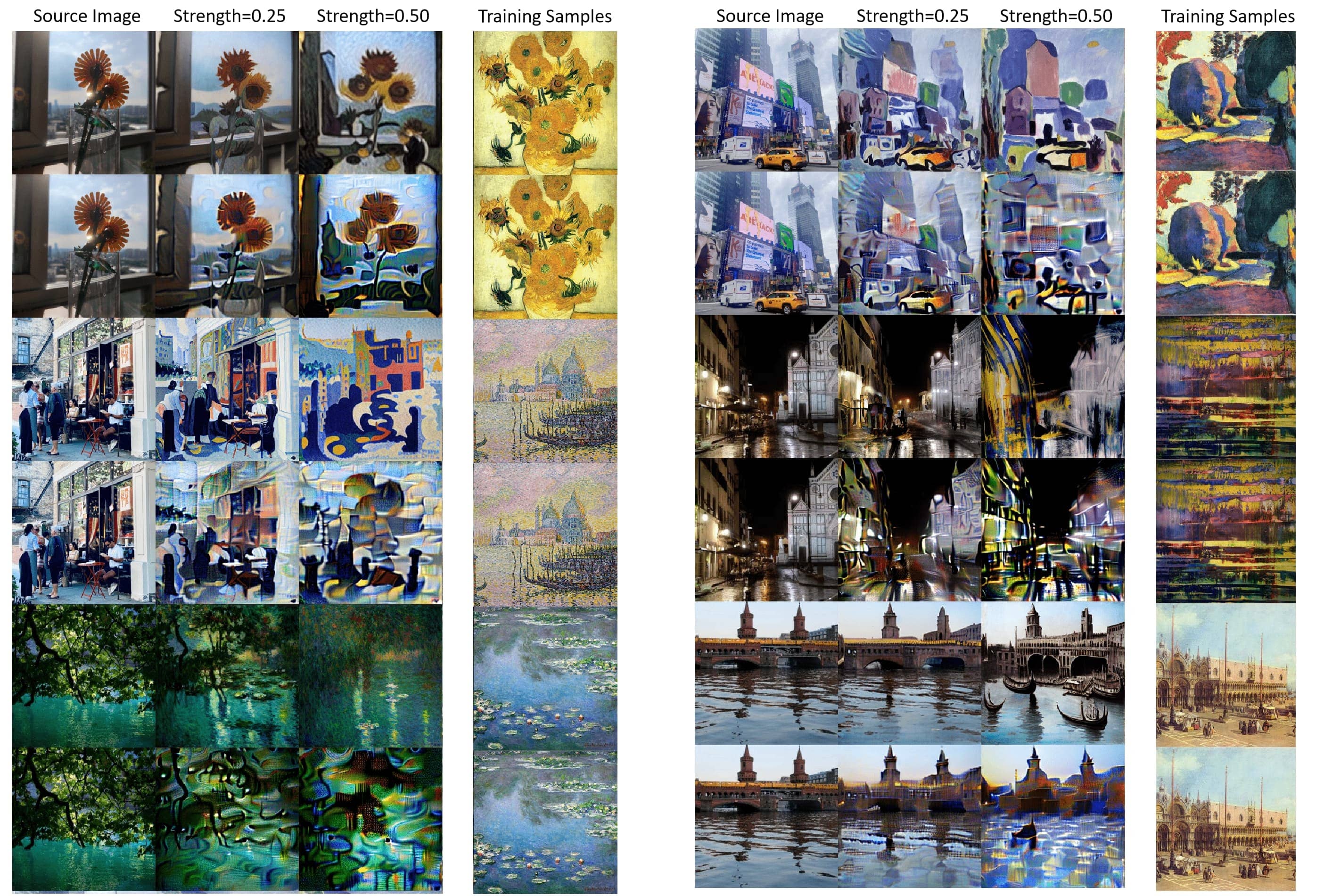}
\caption{Comparison of generated image quality in style transfer for categories of WikiArt~\cite{wikiart}. Images shown in each group share the same source image. We use textual inversion~\cite{gal2022image} to extract the style of training samples from WikiArt, shown in a separate column. For each group, the top row shows the generated images based on the style extracted from the clean examples. The bottom row shows the generated images based on the style extracted from the adversarial examples. Strength is a hyper-parameter that indicates how much the style of the source image is covered by the target style. LDM fails to capture the style from adversarial examples, compared to clean images. }
\label{fig:wikiart}
\end{center}
\vskip -0.2in
\end{figure*}

\subsection{Qualitative Results on Style Transferring}
\label{sec:style_transfer}


An important evaluation scenario for our adversarial examples is the style transfer with LDM, where several copyright violations have taken place~\cite{mimicnews,kimjungginews}. In these cases, infringers first used several image samples to train a pseudo word $S^{*}$ by textual inversion~\cite{gal2022image}, as mentioned in Section~\ref{sam}. Then, they exploited $S^{*}$ in the image-to-image conditional generation and generated images that imitated the art style of the sample images. 

To evaluate the performance of our adversarial examples in resisting this style transfer, we follow this scenario and compare the sample quality of conditionally-generated images based on clean images and adversarial examples. We select 20 paintings of 10 artists respectively from the WikiArt dataset and train an $S^{*}$ for each artist. Other settings are the same as the setting in Section
~\ref{sam}.  As displayed in Figure \ref{fig:wikiart}, the results demonstrate that the style of the conditionally-generated images is significantly different from the input images when conditioning on $S^{*}$ training on adversarial examples. This suggests that AdvDM can be effectively used for copyright protection against illegal style transfer. We further conduct experiments to investigate if our adversarial examples work in Stable Diffusion, a commercialized AI-for-Art application. The results are demonstrated in Appendix~\ref{sec:stable-diffusion}.


\subsection{Qualitative results on image-to-image synthesis}
\label{malicious tampering}

\begin{figure}[ht]
\vspace{-0.1cm}
\begin{center}
\includegraphics[width=0.48\textwidth]{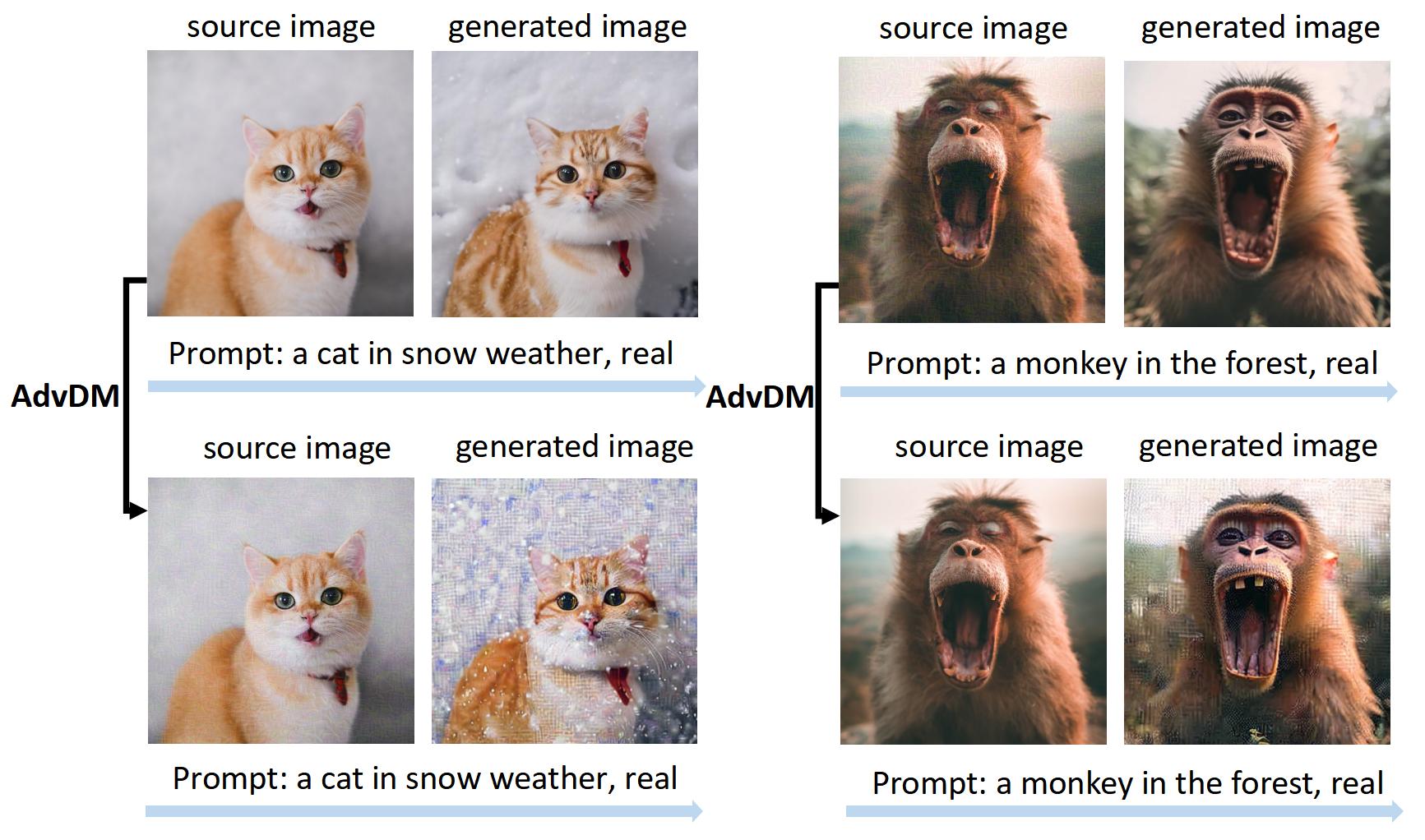}
\vspace{-0.2cm}
\caption{Comparison of images conditionally generated in the image-to-image generation. With conditions extracted from our adversarial examples, LDM generates unrealistic images.}
\label{fig:malicious tampering}
\end{center}
\vspace{-0.3cm}
\end{figure}

As mentioned in Section \ref{sec:3.3}, image-to-image generation is another scenario that measures the quality for adversarial examples. We first apply AdvDM on several open-source photos from Pexels\footnote{https://www.pexels.com/} to generate adversarial examples. Then we generate images based on both these adversarial examples and clean images with the image-to-image pipeline provided by Stable Diffusion, a large-scale commercialized LDM\footnote{https://github.com/huggingface/diffusers}. We compare the quality of generated images in Figure \ref{fig:malicious tampering}. The generated images based on adversarial examples are unrealistic in comparison with those based on clean images.

\subsection{Ablation Study}
\label{ablation stu}
\begin{figure}[t]

 \begin{center}
    \centering
    \subfigure[FID-Sampling Steps]{
        \includegraphics[width=0.48\linewidth]{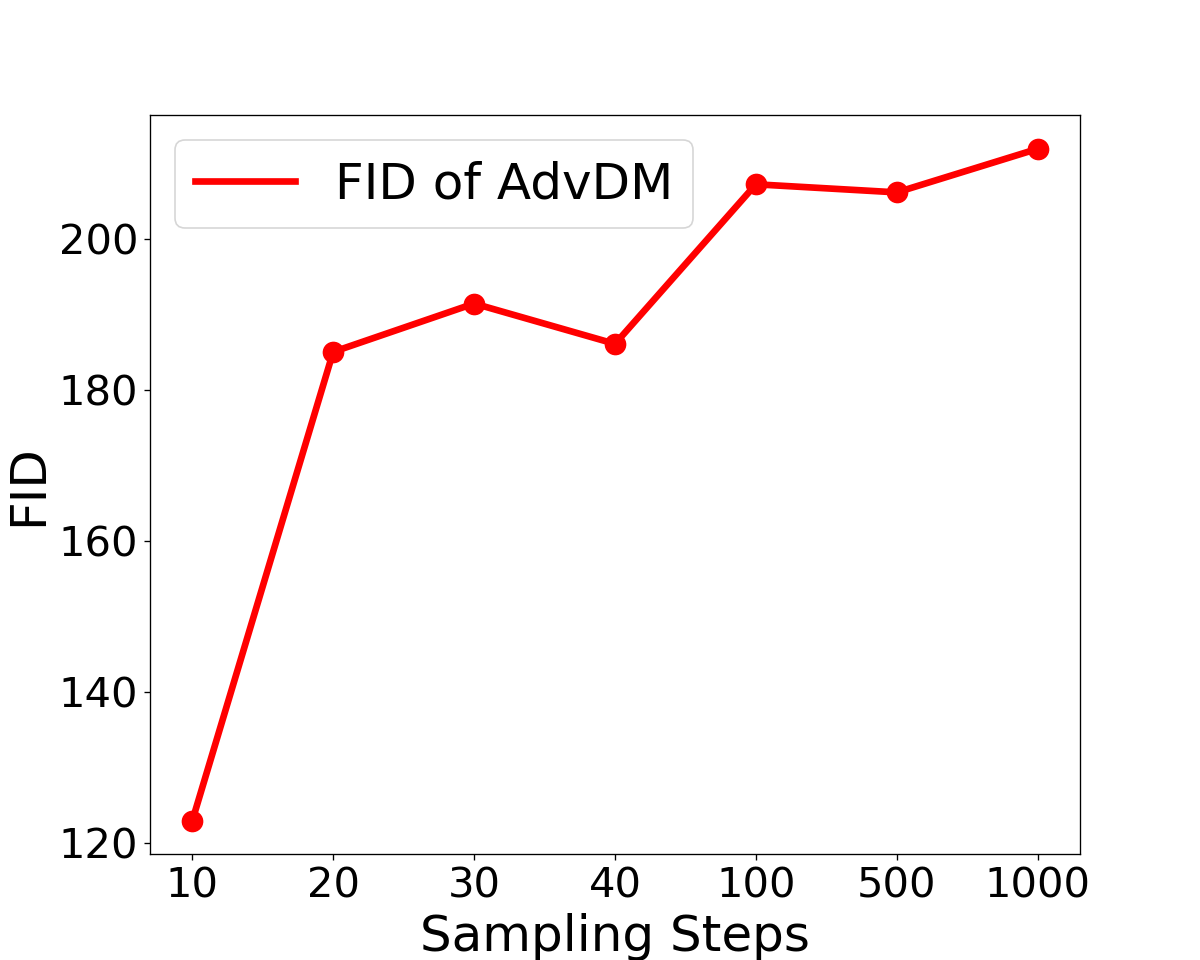}}
    \subfigure[Prec-Sampling Steps]{
        \includegraphics[width=0.48\linewidth]{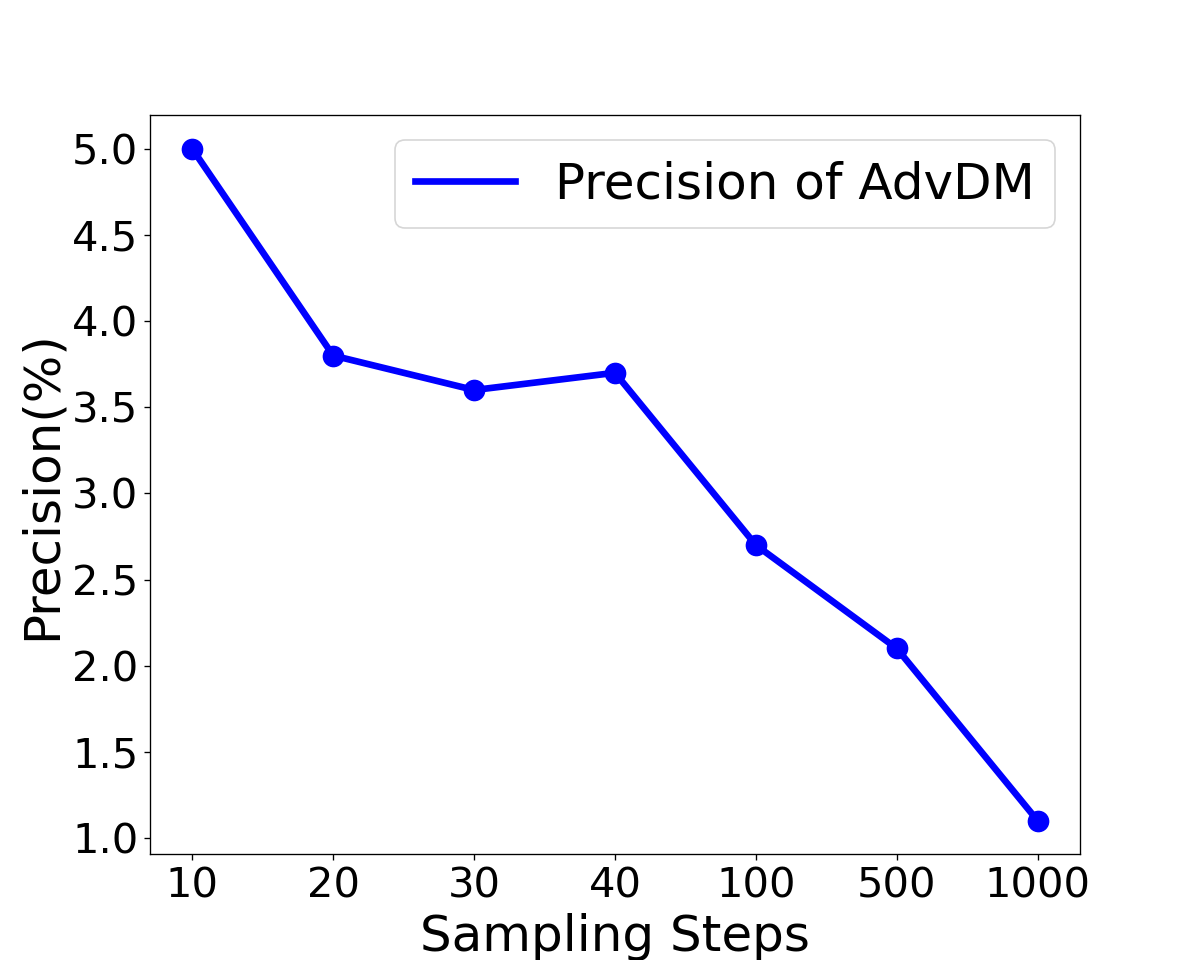}}

    \caption{(a) The FID and sampling steps for AdvDM. (b) The Precision and sampling steps for AdvDM.}

    \label{sampling steps}
\vspace{-0.2cm}
\end{center}
\vspace{-0.3cm}
\end{figure}

\textbf{Sampling steps.}
The number of sampling steps in Monte Carlo is crucial for the accuracy of estimation and thus has a significant impact on the adversarial example generated by AdvDM theoretically. To investigate the effect of this hyper-parameter, we conduct an experiment on the LSUN-airplane dataset, where we pick 100 random images and generate 1,000 images in the setting described in Section~\ref{sam} except for the sampling steps. The number of sampling steps varies from 10 to 1,000. The results are shown in Figure \ref{sampling steps}. With the increase of the sampling steps, the FID increases, and the Precision decreases roughly. It shows that the quality of adversarial examples grows better with more sampling steps.  

It appears that a larger number of sampling steps results in stronger effects on the attack, but also induces inflation in the inference time, as demonstrated in Table \ref{cost}. To balance the tradeoff between performance and inference time, we fix the default sampling step to 40 in our main experiments.

\begin{table}[t]
\setlength\tabcolsep{2.4pt}
\caption{Comparison for AdvDM under different sampling steps. The inference time is the average time to generate an adversarial example over 1,000 images on an NVIDIA RTX A4000 GPU. The unit is second.}
\vspace{0.2cm}
\label{cost}
\begin{center}
\begin{small}
\begin{sc}

\begin{tabular}{cccc}
\hline

\multicolumn{1}{l}{}    Sampling Steps   & FID$\uparrow$ & $prec.\downarrow$ & Inference Time \\ \hline
10  & 122.9        & 0.05             &        1.803      \\
40  & 186.05        & 0.037             &       6.342        \\
1000 & 211.88        & 0.011             &       166.6                 \\ \hline
\end{tabular}

\end{sc}
\end{small}
\end{center}
\vskip -0.2in
\end{table}

\textbf{Perturbation budget.} We also study the impact of the perturbation budget on the quality of the adversarial example generated by AdvDM. We also follow the setting in Section~\ref{sam} except for the perturbation budget. The perturbation budget $\epsilon$ is varied from 2/255 to 32/255. The results are shown in Table \ref{budgets}. We observe that 
 with a small perturbation budget (4/255), AdvDM can already significantly affect the quality of generated images. The visualization results are shown in the Appendix~\ref{vis:ablation}.

\begin{table}[t]
\setlength\tabcolsep{2.4pt}
\caption{The effects of AdvDM under different  perturbation budgets in text-to-image generation scenario}
\label{budgets}
\begin{center}
\begin{small}
\begin{sc}

\begin{tabular}{cccc}
\hline
\multicolumn{1}{l}{}       & \multicolumn{3}{c}{METRIC}                              \\
LSUN airplane       & FID$\uparrow$ & $prec.\downarrow$ & $recall$ \\ \hline
No Attack                  & 54.03         & 0.659             & 0.242               \\
$\epsilon$=2  & 54.49         & 0.295             & 0.276               \\
$\epsilon$=4  & 116.79        & 0.09              & 0.342               \\
$\epsilon$=8  & 186.05        & 0.037             & 0.464               \\
$\epsilon$=16 & 217.09        & 0.015             & 0.569               \\
$\epsilon$=32 & 240.30        & 0.001             & 0.801               \\ \hline
\end{tabular}

\end{sc}
\end{small}
\end{center}
\vskip -0.2in
\end{table}

\subsection{AdvDM vs. Preprocessing Adversarial Defenses}
\label{defense}
There is no existing research that specifically discussed the issue of adversarial defense for diffusion models. One potential approach to defending against AdvDM is by exploiting the use of preprocessing adversarial defenses, which focus on eliminating the adversarial perturbations. This is because they do not ask to retrain the generative model or change the architecture of the model. In light of this, we apply JPEG compression~\cite{das2018shield}, TVM ~\cite{guo2017countering}, and SR ~\cite{mustafa2019image} on adversarial examples generated by AdvDM. The experimental setting about AdvDM follows the same in Section~\ref{sam}.

\begin{figure}[ht]
\vspace{-0.2cm}
\begin{center}
\includegraphics[width=0.48\textwidth]{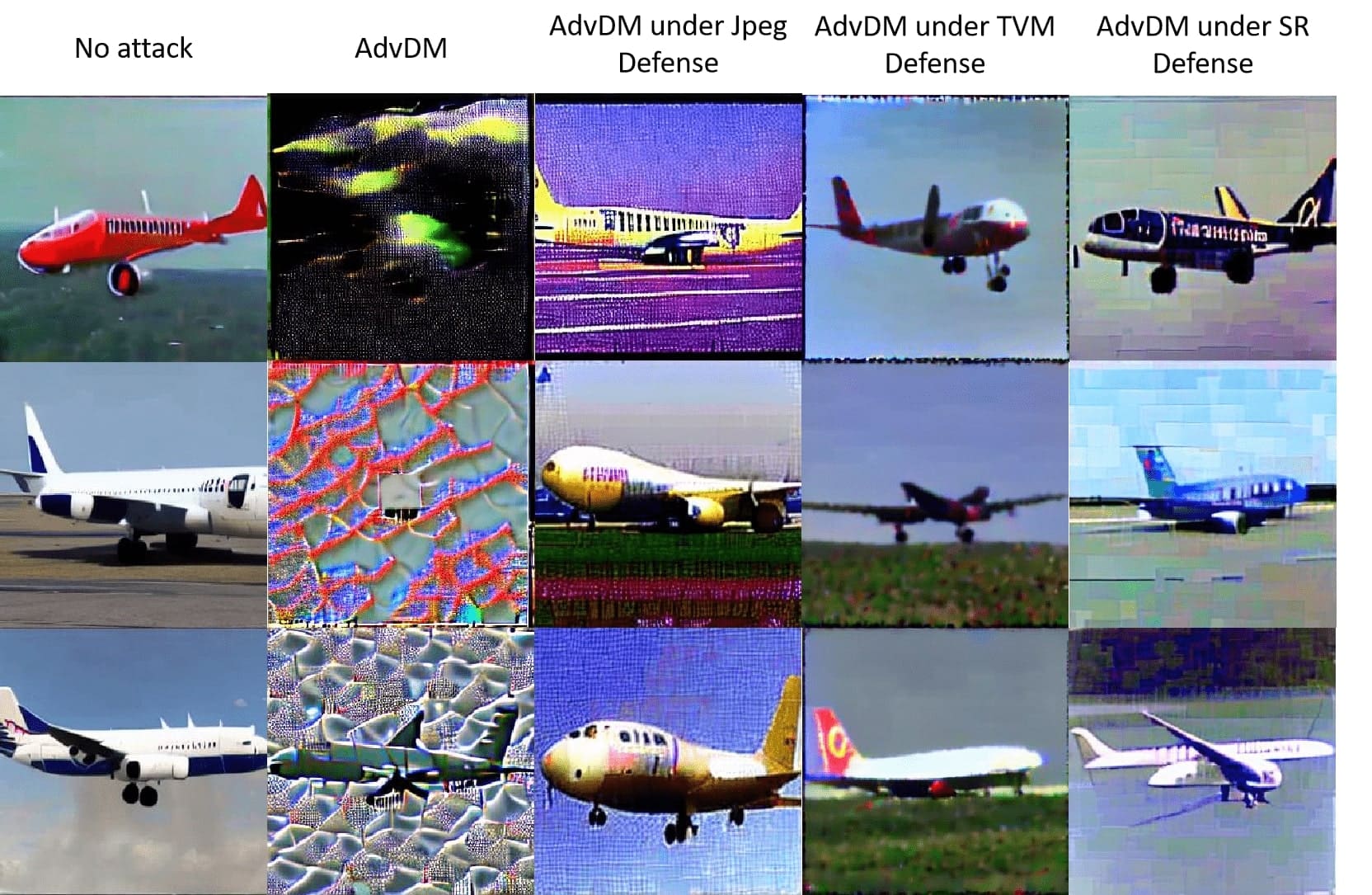}
\caption{Visualization of conditionally-generated images based on different training images. All defenses cannot perfectly maintain the image quality under AdvDM.}
\label{vis:sr}
\vspace{-0.2cm}
\end{center}
\vspace{-0.3cm}
\end{figure}

 The results of AdvDM under preprocessing adversarial defenses are summarized in Table \ref{tab:defense}. It can be observed that both JPEG and TVM have limited effectiveness against the AdvDM attack. SR shows stronger performance in defending, particularly reflected in FID. However, for the Precision, the effectiveness is not significant. This suggests that while preprocessing defenses can partially defend against AdvDM, they are disabled from fully restoring the semantic information of the original images. Furthermore, the differences between images generated from adversarial examples and clean examples are significant, as shown in Figure \ref{vis:sr}.

 Despite the above results, we also apply DiffPure, a state-of-the-art purification-based adversarial defense, to evaluate the robustness of AdvDM. The experiment is demonstrated in Appendix~\ref{appendix:diffpure}.

\begin{table}[htbp]

\setlength\tabcolsep{2.4pt}
\caption{Text-to-image generation based on textual inversion with pre-processing-based adversarial defense}
\label{tab:defense}
\vskip 0.1in
\begin{center}
\begin{small}
\begin{sc}
\resizebox{1\linewidth}{!}{
\begin{tabular}{lcccccc}
\toprule
Defense & \multicolumn{3}{c}{No Defense} &  \multicolumn{3}{c}{JPEG}\\
 Metric & FID $\uparrow$ & $prec.\downarrow$ & $recall$ & FID $\uparrow$ & $prec.\downarrow$ & $recall$\\
\midrule
No attack  &  39.22 & 0.5016 & 0.2765  & 39.19  & 0.5098 & 0.2639 \\
AdvDM   &  169.67 & 0.0263 & 0.3235 & 61.67 & 0.1046 & 0.3208 \\
\midrule
\midrule
Defense & \multicolumn{3}{c}{TVM} &  \multicolumn{3}{c}{SR}\\
 Metric & FID $\uparrow$ & $prec.\downarrow$ & $recall$ & FID $\uparrow$ & $prec.\downarrow$ & $recall$\\
\midrule
No attack  &  44.21 & 0.2513 & 0.1766 & 32.67 & 0.3397 & 0.2332 \\
AdvDM   &  50.95 & 0.1744 & 0.2065  & 40.88  & 0.1673 & 0.2360 \\
\bottomrule
\end{tabular}
}
\end{sc}
\end{small}
\end{center}
\vskip -0.2in
\end{table}


\section{Related Work}
\label{sec:related work}

Adversarial examples have long been an essential topic in different scenarios, including the classification of images~\cite{fgsm} and graphs~\cite{graph,zugner2018adversarial}, text comprehension~\cite{nlp}, and decision making~\cite{rl}. Our definition of the adversarial example for generative models is inspired by that in image classification~\cite{cw}.

Existing research has explored the adversarial example for different generative models yet no proper frameworks have been formulated. Diffusion models are used to improve the adversarial robustness of classifiers~\cite{nie2022diffusion}. Kos \emph{et. al.} studied how to make generative models generate images that would be wrongly classified~\cite{kos2018adversarial}. A theory of adversarial examples for linear flow-based models~\cite{dinh2014nice,dinh2016density,kingma2018glow} has been proposed yet held based on a strong assumption that the data distributes normally, which is not realistic~\cite{pope2020adversarial}. Another study exploited a surrogate attack on classifiers~\cite{fetaya2019understanding}, which is compared with our method in Appendix~\ref{attacks}.



\section{Conclusion}
\label{sec:conclusioin}


In this paper, we are the first to explore and present a theoretical framework to define adversarial examples in diffusion models in order to protect human-created artworks. Based on the framework, we propose an algorithm to generate adversarial examples for diffusion models. Extensive experiments demonstrate that our work provides a paradigm for copyright protection against generative AI and a powerful tool for human artists to protect their artworks from being used without authorization by Diffusion Models-based AI-for-Art applications.


\section*{Acknowledgements}
This research was partly supported by the National NSF of China (NO. 61872234, 61732010), the Shanghai Key Laboratory of Scalable Computing and Systems, and Intel Corporation (UFunding 12679). We extend our heartfelt gratitude to Yichuan Mo and Qingsi Lai from Peking University for their invaluable review and feedback.

Contribution: Chumeng Liang and Xiaoyu Wu are both co-first authors and have made equal contributions to this article. The problem in this paper was initially proposed by Chumeng Liang and Yiming Xue, and refined by Xiaoyu Wu. The algorithm was designed by Xiaoyu and Chumeng. Based on the algorithm, Chumeng formulated the theoretical framework. Xiaoyu then designed and conducted the experiments to evaluate the algorithm. 


\bibliography{example_paper}
\bibliographystyle{icml2023}

\newpage
\appendix
\section{Ethical Issues}
\label{sec:ethicissues}

In this section, we would like to discuss some ethical issues about state-of-the-art AI-for-Art applications based on generative AI and what role our work is expected to play in these issues. 

AI-for-Art applications powered by diffusion models have reshaped the art market by significantly lowering the threshold for artistic creation. However, hidden behind such progress are unresolved copyright issues.

Using copyright-protected training data without the consent of image owners may constitute unauthorized reproduction and distribution, thereby giving rise to copyright infringement liability. One primary source of training data is LAION~\cite{schuhmann2021laion,schuhmann2022laion}, a large-scale dataset of training images with text captions. A large portion of images in LAION was scraped from commercial image-hosting websites without the consent of the image owners~\cite{butterrick2022complaint}. The same issues exist in other generating processes involving unlicensed artworks, for example, learning paintings of a particular artist based on functions of AI-for-Art on a smaller scale without authorization.


Copyright law protects authors’ exclusive rights to reproduce, distribute, perform and display the artworks~\cite{franceschelli2022copyright}. This legal structure makes it highly possible to constitute infringement by using others' artwork without a copyright license in the digital age~\cite{sullivan1996copyright}. Throughout the AI-for-Art process, the transfer of unauthorized artworks from the platform on which it was originally published to AI's database along with the sale or distribution of the program including such database may constitute reproduction and distribution of the original artwork. This is related to the mechanism AI-for-Art applications created artworks. AI-for-Art applications work by fitting the training images and in turn recombining the learned data to generate new images, which may be understood as a special kind of \textit{reproduction}. For some artworks with distinct well-known features, for example, cartoon figures owned by Disney, this reproduction is easy to detect~\cite{disney}. For this reason, the plaintiff lawyer representing artists whose works were used to train these generative AI tools referred to Diffusion Models as “21st-century collage tools” in the recent lawsuit against several companies profiting from Stable Diffusion~\cite{butterrick2022complaint}.

A possible justification for AI-for-Art applications on these issues is the Fair Use Doctrine~\cite{fisher1987reconstructing}. Examples of fair use include criticism, comment, news reporting, teaching, scholarship, and research. It is very likely that training AI with copyright-protected images constitutes fair use for scientific research purposes. However, the generating part is not. It is for commercial purposes and has created millions of dollars for those companies. More than that, AI-for-Art applications compete directly with the artists as a substitute, from whom it obtained its training data. All these facts are disadvantageous to the recognition of fair use.

Copyright law is about a balance between the interests of different participants~\cite{aufderheide2018reclaiming}, as well as the prospect of human creativity. On one side, researchers have made great efforts to develop AI for Art. Such technology revolutionized the method of artistic expression. On the other side, artists are falling behind for lower speed for production and a far higher cost. It takes time for the law to react to new issues brought about by the development of technology, and we look forward to the court’s answers on how the balance will be achieved. But before that, the reality is now severely one-sided – the tech companies make huge money at no cost by appropriating others’ intellectual property, while the artists are left to witness the skills they rely on to make a living being significantly devalued by their own works. The method of protection that this paper proposes aims to arm artists with a weapon to legally protect their statutory rights under copyright law. After all, AI needs to be fair for everyone.
\section{More Visualization}
\label{visualization}
\subsection{Ablation Study}
\label{vis:ablation}
\begin{figure}[ht]

\begin{center}
\includegraphics[width=0.49\textwidth]{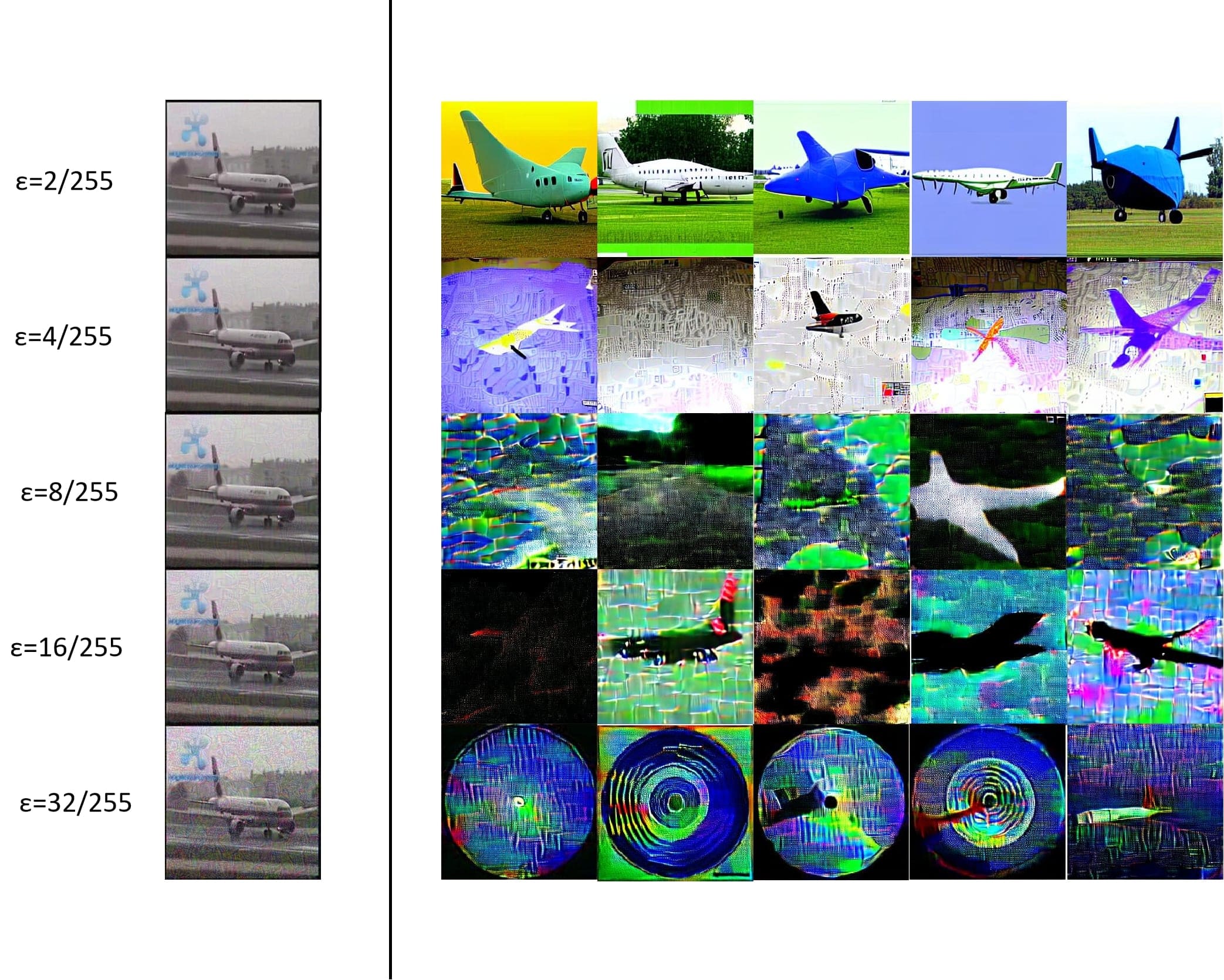}
\caption{Visualization of ablation study in perturbation budgets. First column: adversarial examples. The second to the sixth column: the image generated conditioned on the pseudo-word derived from adversarial examples for different samplings.}
\label{vis_epi}
\end{center}
\end{figure}

We provide visualization of the generated images based on adversarial examples under different perturbation budgets in Figure \ref{vis_epi}. With a greater perturbation budget, the figure of airplanes grows vaguer.

\subsection{Text-to-image generation based on textual inversion}
\label{vis-txt2img}
\begin{figure}[ht]
\vskip 0.2in
\begin{center}
\includegraphics[width=0.47\textwidth]{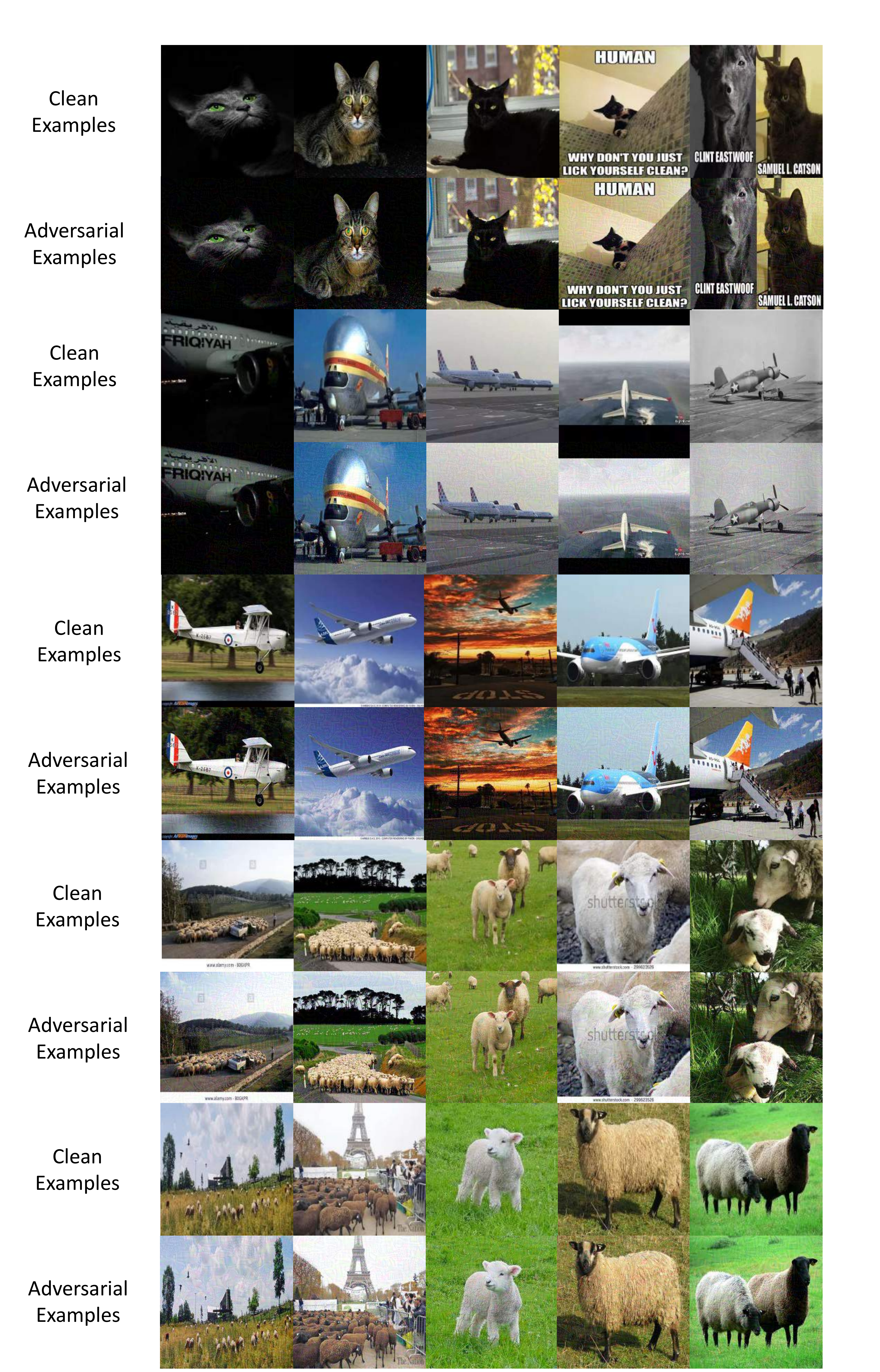}
\caption{One group of the clean images and adversarial images in Lsun-cat, Lsun-sheep, Lsun-airplane dataset. The top row shows the clean examples while the bottom row shows the adversarial examples under AdvDM.}
\label{merge_source}
\end{center}
\vskip -0.2in
\end{figure}

\begin{figure*}[htbp]
\vskip 0.2in
\begin{center}
\includegraphics[width=0.98\textwidth]{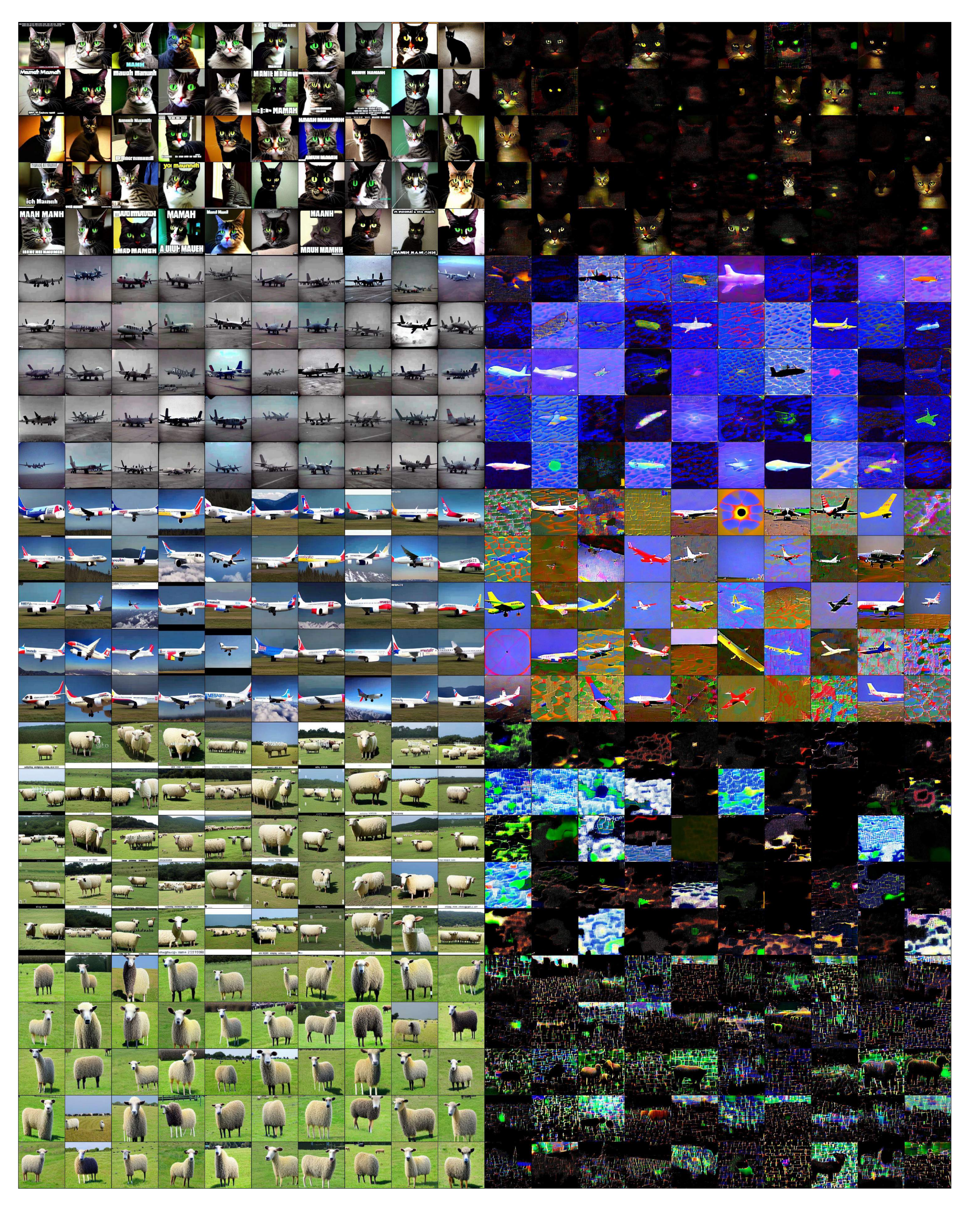}
\caption{Left: Text-to-image generation based on textual inversion for clean examples. Right: Text-to-image generation based on textual inversion for adversarial examples.}
\label{mergegen}
\end{center}
\vskip -0.2in
\end{figure*}

We compare the adversarial examples with the clean images they are generated from in Figure~\ref{merge_source}. There are almost no human-visible differences between adversarial examples and clean images. We then generate images with these adversarial examples and clean images by text-to-image generation. The results are shown in Figure \ref{mergegen} and indicates that our adversarial examples severely decrease the quality of generated images.

\section{Comparison with Other Potential Adversarial Examples}
\label{attacks}
There is no previous research on adversarial examples for diffusion models. Therefore, in addition to AdvDM, we also investigate several other potential adversarial examples for diffusion models for a complete understanding of AdvDM. Note that there are no formulated methods to generate adversarial examples for diffusion models. We then explore more potential methods inspired by existing research.

For the experimental setting, the steps and perturbation budgets for these adversarial examples are constrained under the same settings as outlined in Section \ref{ablation stu}. 
\subsection{PGD on Classifiers}

 It is shown that some conditional generative models are vulnerable even to some adversarial examples generated for classification models~\cite{fetaya2019understanding}. Also, the transferability of adversarial examples between neural networks is widely validated and exploited ~\cite{papernot2016transferability,zugner2018adversarial}. Following this idea, we generate adversarial examples by Projected Gradient Descent~\cite{pgd} (PGD) on an InceptionV3 classifier~\cite{xia2017inception}. We then consider these adversarial examples to be transferable adversarial examples for LDM. The method is denoted by \textbf{PGD (InceptionV3)}.

\subsection{Attacking the Embedding Layer}
Note that LDM includes an embedding layer that projects images to a representation in the latent space. This can be regarded as an encoder-decoder structure in AutoEncoder~\cite{rumelhart1985learning}. It is shown by existing research that the encoder-decoder structure can be exploited to generate adversarial examples~\cite{kos2018adversarial}. Inspired by this idea, we apply PGD~\cite{pgd} to the embedding layer. We compute a new term of loss by comparing the latent representation of the clean image and that of the adversarial example, which is obtained by adding a tiny perturbation $\delta$ to the clean image. The optimization goal is to maximize the loss by the perturbation. We denote this method by \textbf{Embedding Attack}, for it generates adversarial examples by applying an adversarial attack against the embedding layer in LDM.
\begin{definition}[\textbf{Adversarial Example for Diffusion Models (with Embedding Attack)}]
 Denote the encoder in the LDM by $\mathcal{E}$. $x$ is the input image and $\delta$ is the perturbation under a certain budget. The adversarial example generated by Embedding Attack is formulated as $x':=x+\delta$, where

    \begin{equation}
       \begin{aligned}
       \delta:=&\arg\max\limits_{\delta}\mathcal{L}_{embedding}(x, \delta)\\ =& \arg\max\limits_{\delta}\Vert\mathcal{E}(x) - \mathcal{E}(x + \delta)\Vert_2.
       \end{aligned}
    \end{equation}

\end{definition}

We denote the adversarial example in the optimization step $i$ by $x^{(i)}$. For implementation, we follow the default setting of PGD~\cite{pgd} and randomly initialize the perturbation at the beginning of the optimization by $x^{(0)} = x + \epsilon z$, where $z \in \mathcal{N}(0,1)$ and $\epsilon$ is the perturbation budget of the attacks. The adversarial examples are crafted by an iterative multi-step signed gradient ascent with step length $\alpha$. The number of iteration steps is set to 40. The optimization process is summarized as

    \begin{equation}
    \label{eq:embedding_attack}
            x^{(i+1)} = x^{(i)} +\alpha\mbox{sgn}(\nabla_{x^{(i)}}\mathcal{L}_{embedding}(x, x^{(i)} - x)),
    \end{equation}
where $\mbox{sgn}$ refers to the sign function.

\subsection{PGD}
Another method to generate adversarial examples is to apply PGD to the loss of LDM. This is equivalent to our method when the number of sampling steps $N$ is 1. The method is denoted by \textbf{PGD (LDM)}.

\begin{table}[htbp]
\setlength\tabcolsep{2.4pt}
\caption{Text-to-image generation based on textual inversion using adversarial examples under different possible attacks}
\label{Tab: Attack}
\vskip 0.15in
\begin{center}
\begin{small}
\begin{sc}

\begin{tabular}{cccc}
\hline
\multicolumn{1}{l}{} & \multicolumn{3}{c}{METRIC}                              \\
\multicolumn{1}{l}{} & FID$\uparrow$ & $prec.\downarrow$ & $recall.$ \\ \hline
No Attack            & 55.19         & 0.547             & 0.231               \\
PGD (InceptionV3)    & 56.89         & 0.306             & 0.153               \\
Embedding Attack     & 175.34        & \textbf{0.023}             & 0.352               \\
PGD (LDM)   & 164.38        & 0.042             & 0.438               \\
AdvDM                & \textbf{186.05}        & 0.037              & 0.464              \\ \hline
\end{tabular}

\end{sc}
\end{small}
\end{center}
\vskip -0.1in
\end{table}

The results of these experiments are presented in Table \ref{Tab: Attack}. As can be observed from the table,  AdvDM achieves the best results among all the methods benchmarked by FID. Embedding attacks also show relatively promising results, especially in Precision. On the other hand, PGD on DMs, which lacks the sampling process, fails to effectively decrease the probability $p_{\theta}$, leading to poorer performance. Classifier attacks, which involve transferring an attack on a classifier to the generation model, do not show much effect, indicating that this method is not directly effective in this setting.

\begin{figure}[ht]

\begin{center}
\includegraphics[width=0.48\textwidth]{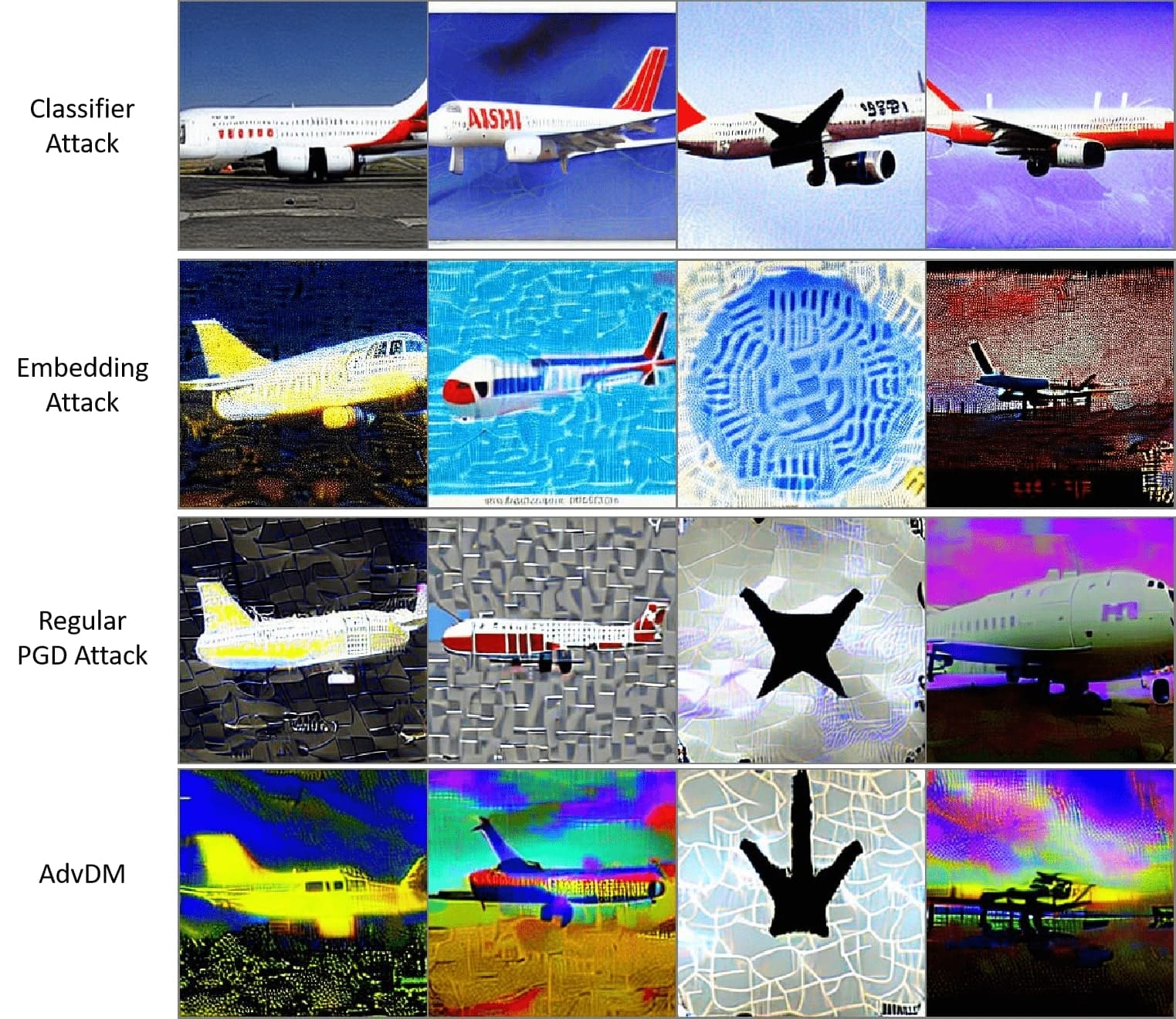}
\caption{Visualization of generated images under different attacks. The pseudo-word used for generation is derived from Lsun-airplane dataset, where we can observe that AdvDM can more effectively influence the semantic information of the images.}
\label{attacks_vis}
\vskip -0.2in
\end{center}
\vskip -0.2in

\end{figure}

We also provide visualization for the generation under different attacks in Figure \ref{attacks_vis}. From the visualization, we observe that under embedding attacks, while noise is created in the background of the generated images, the semantic information of the images is not largely destroyed. However, under AdvDM, the semantic information (such as the shape or the color) of the images is largely affected, indicating a stronger attacking effect.

\section{Implementation Details}
\label{appendix:eva}
\begin{algorithm}[htbp]
   \caption{Text-to-image generation based on textual inversion}
   \label{alg:t2i}
\begin{algorithmic}
   \STATE {\bfseries Input:}  Dataset $\mathcal{D}$, latent diffusion model $\theta$ with latent encoder $\varepsilon_{\theta}$ and embedded language model $\mathcal{F}_{\theta}$, optimization steps $N$, step length $\alpha$
   \STATE Randomly pick 1,000 images $X$ from $\mathcal{D}$
   \STATE Separate $X$ into 5-image groups $X:=\{x^{0}, x^{1},x^{2},...\}$
   \STATE Initialize $X_g\leftarrow\emptyset$
   \FOR{$x^{i}$ {\bfseries in} $X$}
   \STATE Initialize $S^{*}$ with a random word vector in the semantic space of $\mathcal{F}_{\theta}$
   \STATE Optimizing a pseudo word $S^{*}$:
   \FOR{$j=1$ {\bfseries to} $N$}
   \STATE Randomly pick an image $x$ from $x^{i}$
   \STATE $c\leftarrow$ ``A photo of $S^{*}$"
   \STATE Compute $\mathcal{L}_{j}=\mathbb{E}_{z\sim\varepsilon_{\theta}(x^{i}),c, t}||\epsilon_t-\epsilon_{\theta}(z,t,\mathcal{F}_{\theta}(c))||^2_2$
   \STATE $S^{*}\leftarrow S^{*}-\alpha\nabla_{S^{*}}\mathcal{L}_{j}$
   \ENDFOR
   \STATE $c_g\leftarrow$ ``A photo of $S^{*}$"
   \STATE Use $c_g$ to generate 50 images $x_g^{i}$ by the text-to-image function of $\theta$
   \STATE $X_g\leftarrow X_g\cup x_g^{i}$
   \ENDFOR
   \STATE Compute FID, $prec$, and $recall$ between $X_g$ and $\mathcal{D}$
\end{algorithmic}
\end{algorithm}

\begin{algorithm}[htbp]
   \caption{Style transfer based on textual inversion}
   \label{alg:sty}
\begin{algorithmic}
   \STATE {\bfseries Input:} Dataset $\mathcal{D}$, latent diffusion model $\theta$ with latent encoder $\varepsilon_{\theta}$ and embedded language model $\mathcal{F}_{\theta}$, optimization steps $N$, step length $\alpha$
   \STATE Separate $D$ based on different styles: $D^{'}:=\{x^{0}, x^{1},x^{2},...,x^{k}\}$.
   \STATE Initialize $X_g\leftarrow\emptyset$
   \FOR{$x^{i}$ {\bfseries in} $D^{'}$}
   \STATE Initialize $S^{*}$ with a random word vector in the semantic space of $\mathcal{F}_{\theta}$
   \STATE Optimizing a pseudo word $S^{*}$:
   \FOR{$j=1$ {\bfseries to} $N$}
   \STATE Randomly pick an image $x$ from $g^{i}$
   \STATE $c\leftarrow$ ``A painting in the style of $S^{*}$"
   \STATE Compute $\mathcal{L}_{j}=\mathbb{E}_{z\sim\varepsilon_{\theta}(x^{i}),c, t}||\epsilon_t-\epsilon_{\theta}(z,t,\mathcal{F}_{\theta}(c))||^2_2$
   \STATE $S^{*}\leftarrow S^{*}-\alpha\nabla_{S^{*}}\mathcal{L}_{j}$
   \ENDFOR
   \STATE $c_g\leftarrow$ A target prompt containing ``in the style of $S^{*}$"
   \STATE $\overline{x}\leftarrow$ A photo or script
   \STATE Use $c_g$, $\overline{x}$ to generate images $x_g^{i}$ by the image-to-image function of $\theta$
   \STATE $X_g\leftarrow X_g\cup x_g^{i}$
   \ENDFOR
   \STATE \textbf{Return} $X_g$
\end{algorithmic}
\end{algorithm}

\begin{algorithm}[htbp]
   \caption{Image-to-image synthesis}
   \label{alg:img2img}
\begin{algorithmic}
   \STATE {\bfseries Input:} Source Images $X$,  latent diffusion model $\theta$ with latent encoder $\varepsilon_{\theta}$ and embedded language model $\mathcal{F}_{\theta}$
   \STATE Initialize $X_g\leftarrow\emptyset$   
   \FOR{$x$ {\bfseries in} $X$}
   \STATE $c\leftarrow$ A target prompt for image $x$
   \STATE Use $c$, $x$ to generate images $x_g$ by the image-to-image function of $\theta$
   \STATE $X_g\leftarrow X_g\cup x_g$
   \ENDFOR
   \STATE \textbf{Return} $X_g$
\end{algorithmic}
\end{algorithm}
\subsection{Details of Evaluation}
\label{scenario}

We describe the detailed procedure of the three evaluation scenarios in Algorithm~\ref{alg:t2i}, Algorithm~\ref{alg:sty}, and Algorithm~\ref{alg:img2img}, respectively. We use the pre-trained model provided by the author of the latent diffusion model~\cite{rombach2022high}. For text-to-image generation and style transfer, the procedure follows the setting recommended by the textual inversion paper~\cite{gal2022image}. For style transfer, we fix the strength to 0.5. We also follow the paper to choose $N$ as 5000. For image-to-image, we follow the default setting in~\cite{rombach2022high}.

\label{details_eva}
For FID scores, we use an open-source package \footnote{https://github.com/w86763777/pytorch-gan-metrics}. For Precision- and Recall- scores, we use the script provided by Dhariwal and Nichol~\cite{dhariwal2021diffusion}. Three metrics are calculated over the whole category dataset. 

\subsection{Implementation of AdvDM}
\label{details_advdm}
\begin{algorithm}[htbp]
   \caption{Implementation of AdvDM on  Latent Diffusion Models}
   \label{alg:Adv-details-ldm}
\begin{algorithmic}
   \STATE {\bfseries Input:} Data $x_0$, parameter $\theta$,  denoising autoencoder $\varepsilon_{\theta}$, encoder $\mathcal{E}$,  number of Monte Carlo $N$, step-wise perturbation budget $\alpha$, overall perturbation budget $\epsilon$
   \STATE {\bfseries Output:} Adversarial example $x'_0$
   \STATE Initialize $x_0^{(0)} \leftarrow x_0$.

   \FOR{$i=1$ {\bfseries to} $N$}
   \STATE Sample $x^{(i)}_{1:T}\sim q(\mathcal{E}(x^{(i)}_{1:T})|\mathcal{E}(x_0^{(i)}))$
   \STATE Sample $t \sim U(1, T)$

   \STATE $\delta^{(i)}\leftarrow\alpha \mbox{sgn}(\nabla_{x_0^{(i)}} \mathcal\Vert \mathcal{E}(x^{(i)}_{T})- \varepsilon_{\theta}(\mathcal{E}(x^{(i)}_{t}), t)\Vert_{2})$

   \STATE Clip $\delta^{(i)}$ s.t. $\Vert x_0^{(i-1)}+ \delta^{(i)} - x_0^{(0)}\Vert_{\infty}\leq \epsilon$

      \STATE $x_0^{(i)}\leftarrow x_0^{(i-1)}+ \delta^{(i)} $
   \ENDFOR
   \STATE $x'_0\leftarrow x_0^{(N)}$
\end{algorithmic}
\end{algorithm}

We implement AdvDM on Latent Diffusion Models. As shown in Algorithm \ref{alg:Adv-details-ldm}, adversarial perturbation is added to the original image under a sampling series $x^{(i)}_{1:T}$ and a random timestamp $t$ for each step.

\section{Protection Effectiveness against Stable Diffusion}
\label{sec:stable-diffusion}

Our motivation is to protect paintings created by human artists from being imitated by AI-for-Art applications. Note that various mainstream AI-for-Art applications~\footnote{Text\mbox{2}Dream: https://deepdreamgenerator.com/\#tools}~\footnote{Night Cafe: https://creator.nightcafe.studio/stable-diffusion-image-generator}~\footnote{Hotpot: https://hotpot.ai/stable-diffusion}~\footnote{NovelAI: https://novelai.net/} use the model with the architecture of LDM. Hence, we expect satisfying protection effectiveness of our adversarial examples against these AI-for-Art applications. Here, we conduct an experiment to evaluate the protection effectiveness against Stable Diffusion, a famous AI-for-Art application. Note that the model~\footnote{https://huggingface.co/CompVis/stable-diffusion-v-1-4-original} used by Stable Diffusion has a similar architecture as LDM~\footnote{https://ommer-lab.com/files/latent-diffusion/nitro/txt2img-f8-large/model.ckpt} but it has a larger scale with more parameters.

\begin{figure*}[htbp]
\vskip 0.2in
\begin{center}
\includegraphics[width=0.90\textwidth]{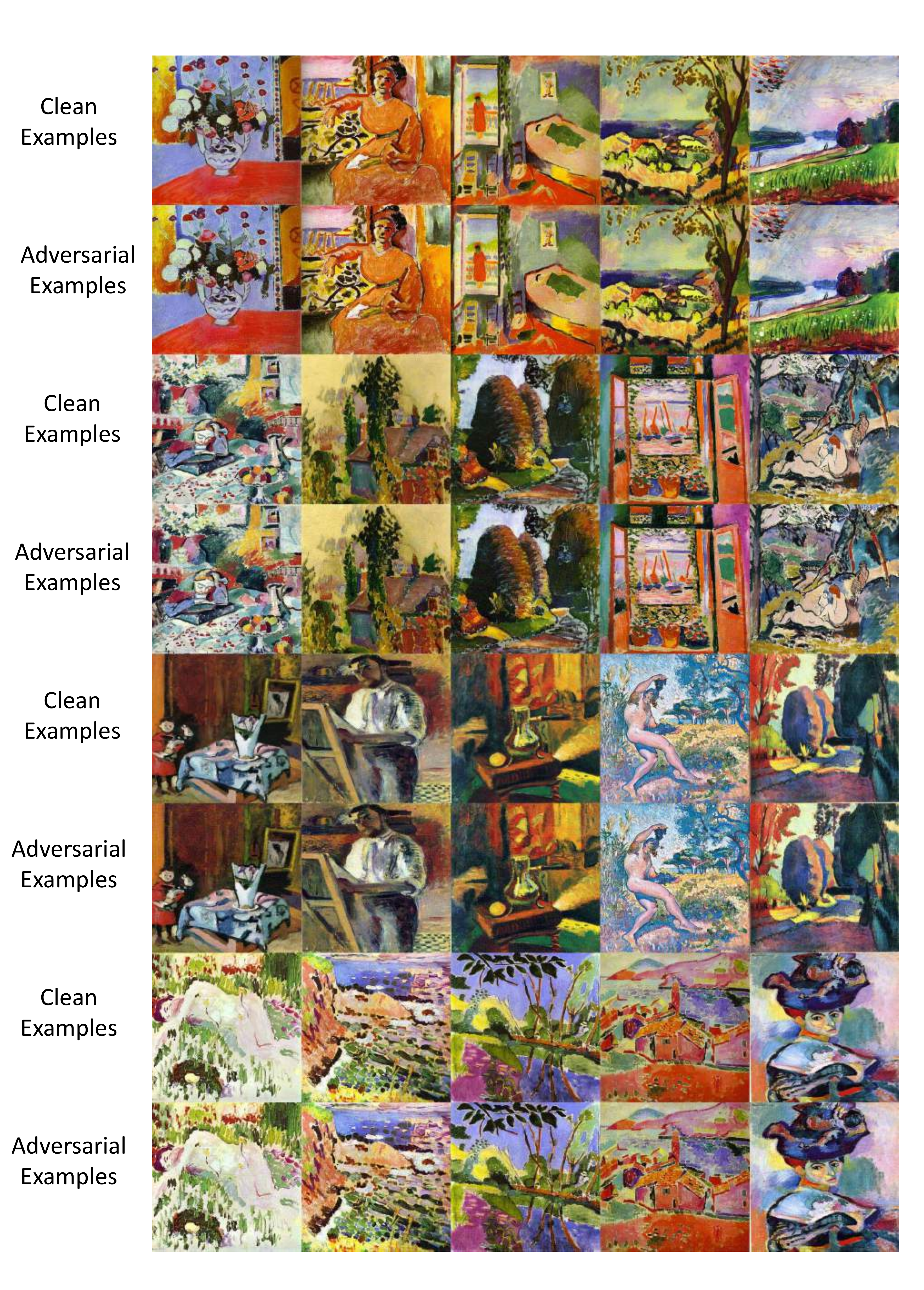}
\caption{Clean examples and adversarial examples of Henri Matisse's paintings.}
\label{matisse}
\end{center}
\vskip -0.2in
\end{figure*}

\begin{figure*}[htbp]
\vskip 0.2in
\begin{center}
\includegraphics[width=0.90\textwidth]{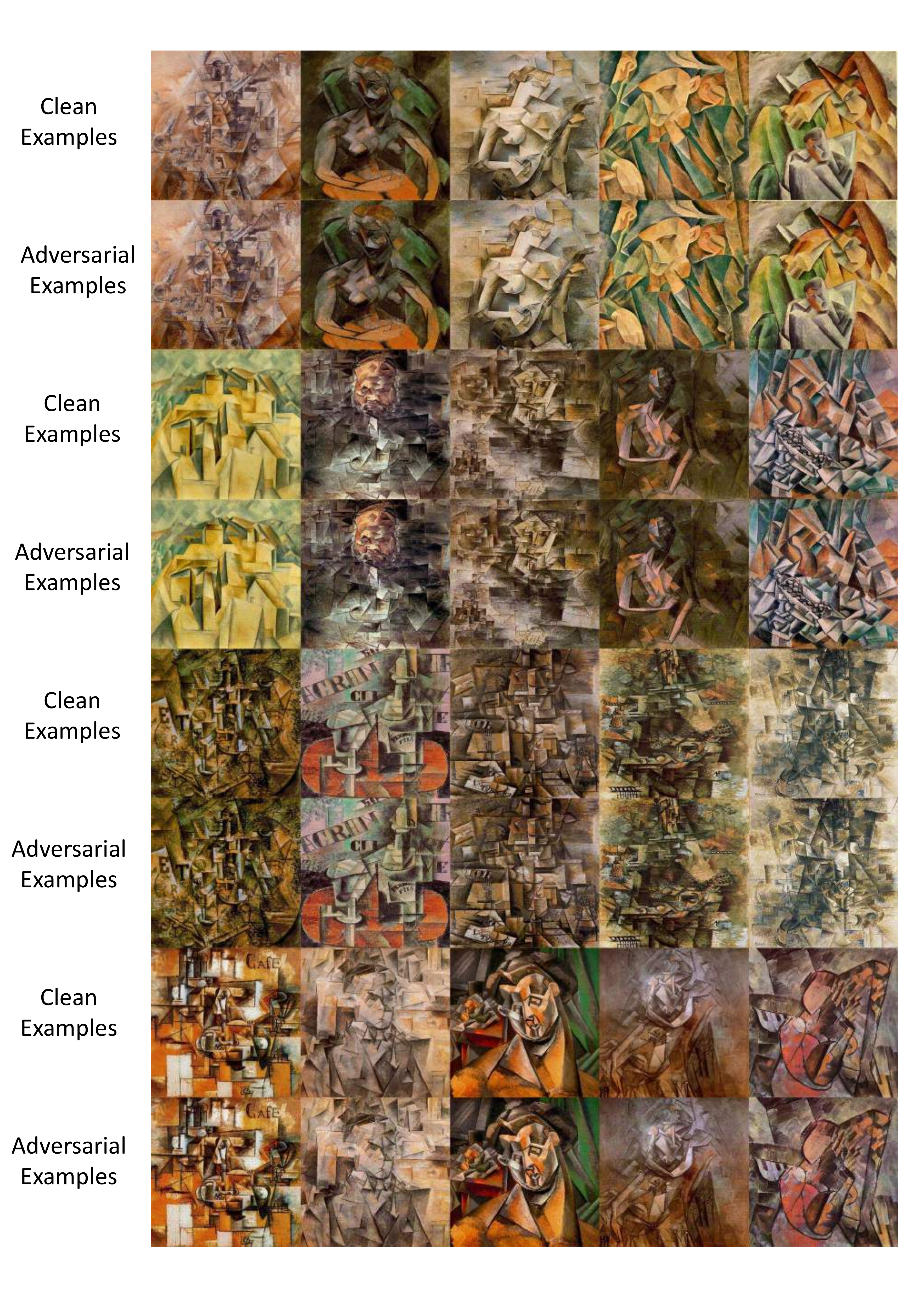}
\caption{Clean examples and adversarial examples of Pablo Picasso's paintings.}
\label{piccaso}
\end{center}
\vskip -0.2in
\end{figure*}

\begin{figure*}[htbp]
\vskip 0.2in
\begin{center}
\includegraphics[width=0.90\textwidth]{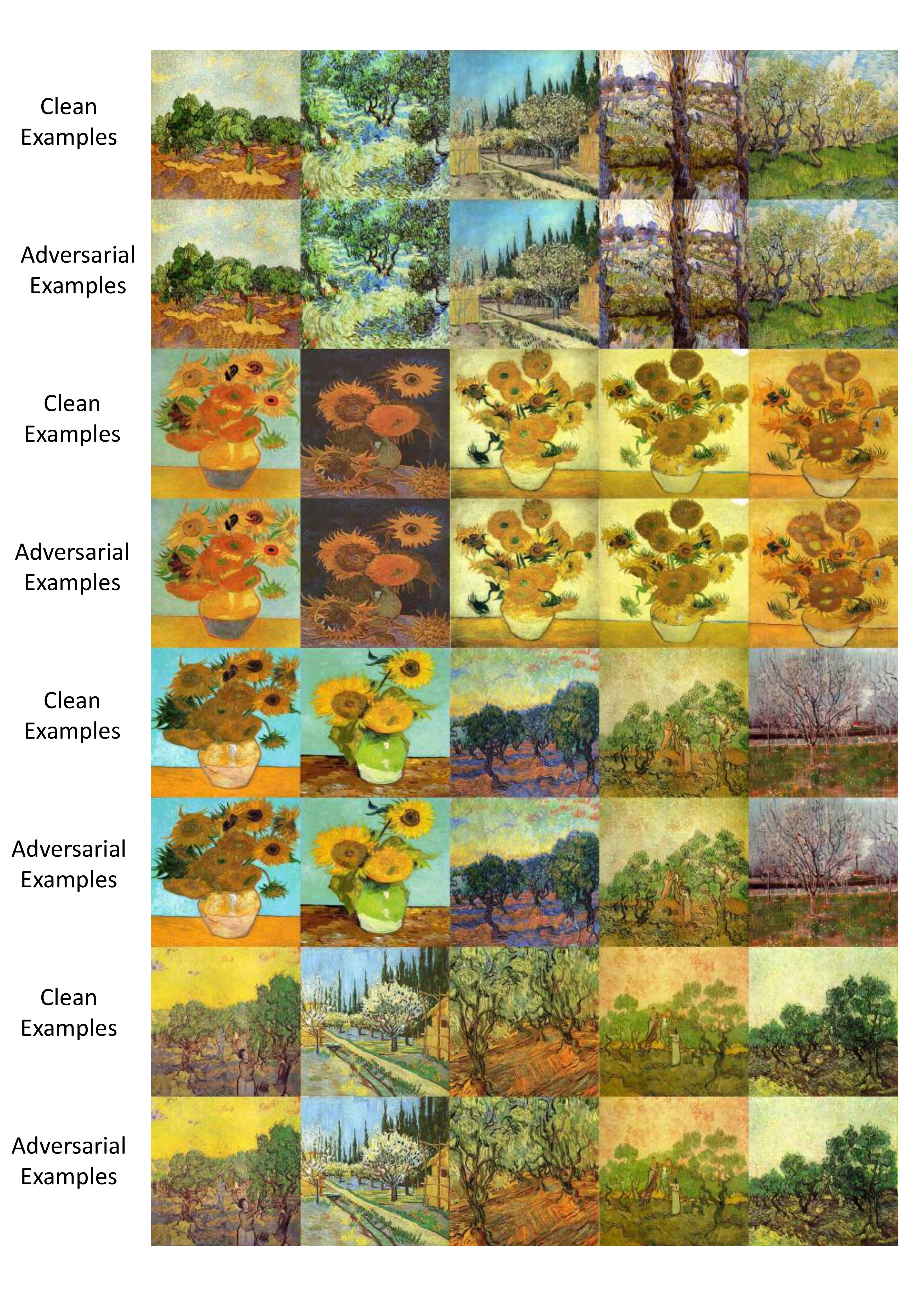}
\caption{Clean examples and adversarial examples of Vincent Van Gogh's paintings.}
\label{vangogh}
\end{center}
\vskip -0.2in
\end{figure*}

\begin{figure*}[htbp]
\vskip 0.2in
\begin{center}
\includegraphics[width=0.90\textwidth]{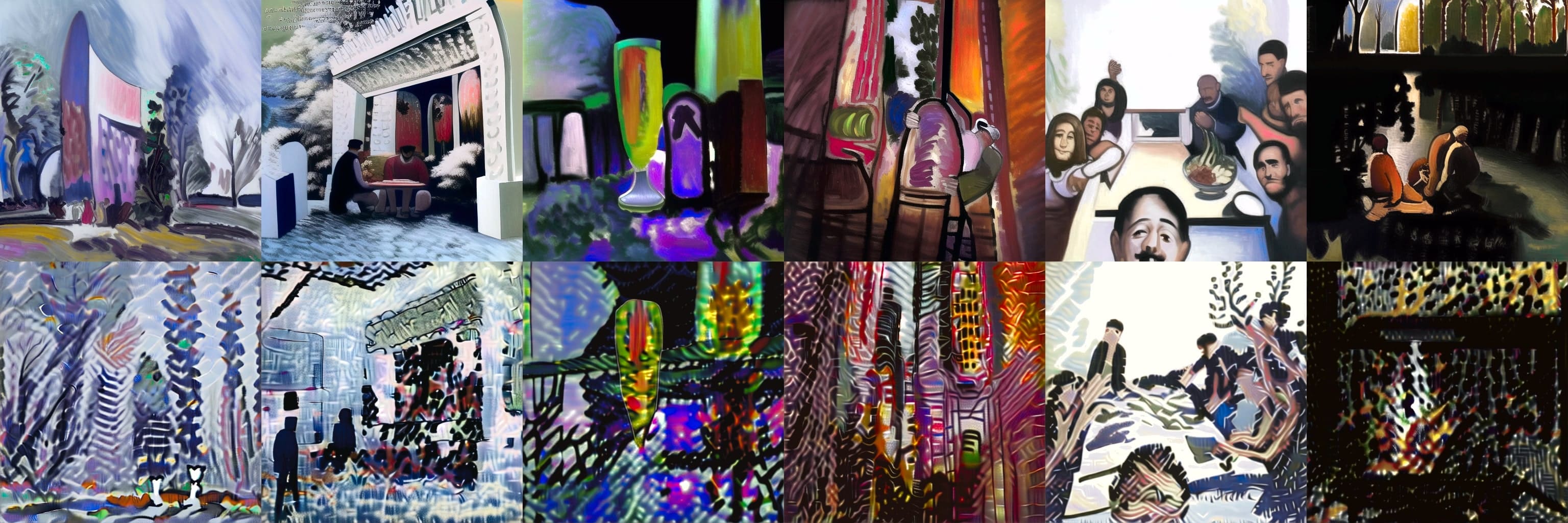}
\caption{
\textbf{The first row:}  Generated images by Stable Diffusion based on clean examples of  Henri Matisse's paintings. \textbf{The second  row:} Generated images by Stable Diffusion based on adversarial examples of Henri Matisse's paintings.}
\label{matisse-gen}
\end{center}
\vskip -0.2in
\end{figure*}

\begin{figure*}[htbp]
\vskip 0.2in
\begin{center}
\includegraphics[width=0.90\textwidth]{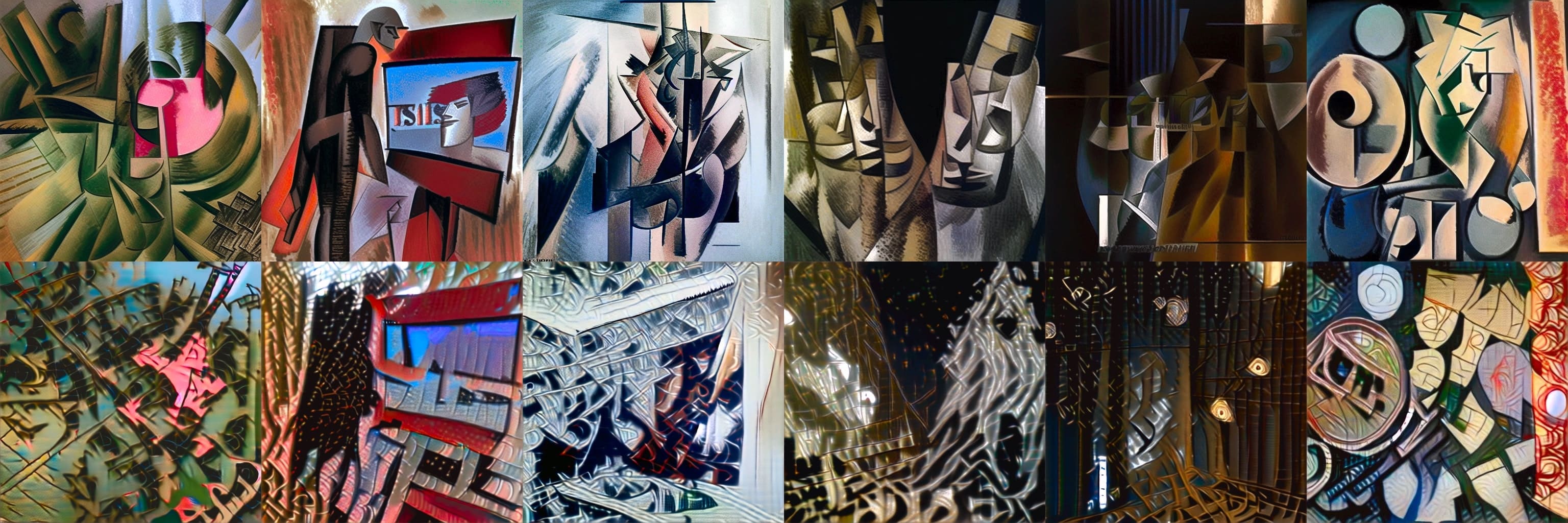}
\caption{\textbf{The first row:} Generated images by Stable Diffusion based on clean examples of Pablo Picasso's paintings. \textbf{The second  row:} Generated images by Stable Diffusion based on adversarial examples of Pablo Picasso's paintings. }

\label{piccaso-gen}
\end{center}
\vskip -0.2in
\end{figure*}

\begin{figure*}[htbp]
\vskip 0.2in
\begin{center}
\includegraphics[width=0.90\textwidth]{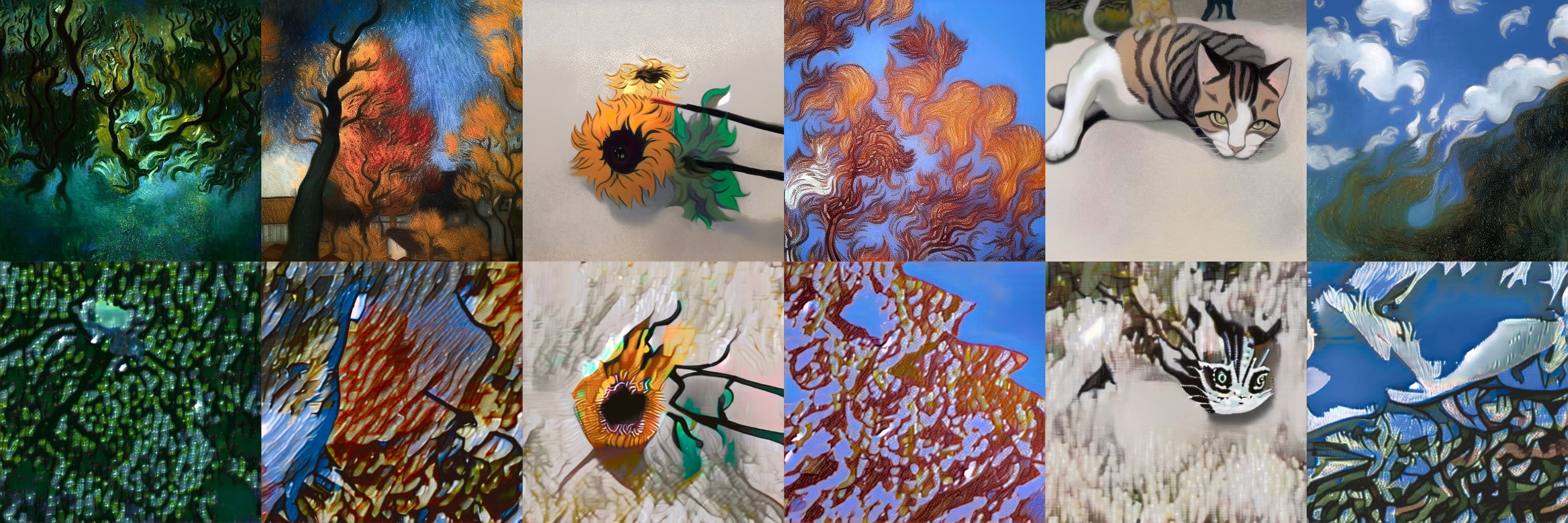}
\caption{\textbf{The first row:} Generated images by Stable Diffusion based on clean examples of  Van Gogh's paintings. \textbf{The second row:}  Generated images by Stable Diffusion based on adversarial examples of Vincent Van Gogh's paintings. }

\label{vangogh-gen}
\end{center}
\vskip -0.2in
\end{figure*}

We evaluate our adversarial examples on the WikiArt dataset~\cite{wikiart}. We select 20 paintings from three artists respectively: Vincent Van Gogh, Pablo Picasso, and Henri Matisse. We then generate adversarial examples based on these paintings. The number of sampling steps $N$ is set to 100. The perturbation budget $\epsilon$ is 8/255 and the step length $\alpha$ is 1/255. Then, we do style transfer with textual inversion on Stable Diffusion. The procedure is very similar to that described by Algorithm~\ref{alg:sty}. For the optimization of the pseudo word $S^{*}$, the optimization step is 8000 with a step length of 0.005. The reconstruction strength is set to 0.5. We first compare the clean paintings used for optimizing $S^{*}$ with the adversarial examples in Figure~\ref{matisse}, Figure~\ref{piccaso}, and Figure~\ref{vangogh}. Then, we visualize the results of generated images on clean paintings and adversarial examples in Figure~\ref{matisse-gen}, Figure~\ref{piccaso-gen}, and Figure~\ref{vangogh-gen}.

\newpage
\section{Additional Experiments}
\label{appendix:additional}
\begin{figure*}[tbp]
\begin{center}
\includegraphics[width=0.9\textwidth]{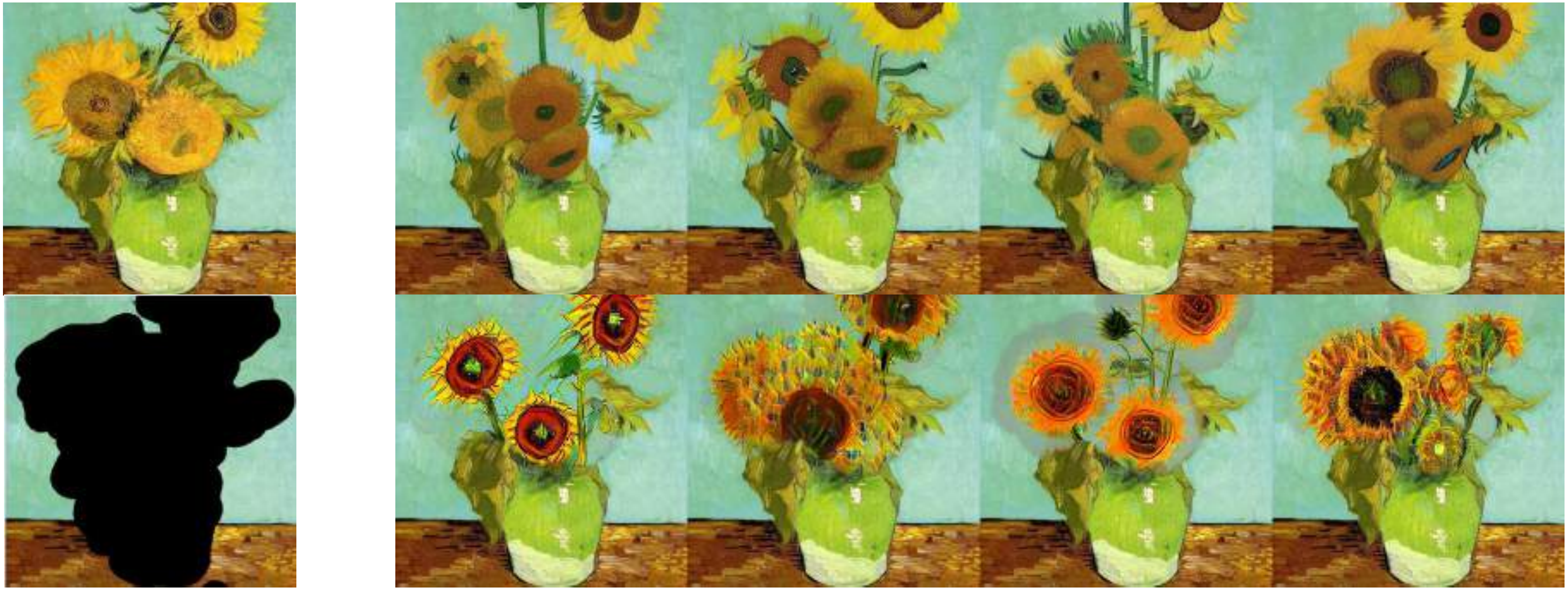}
\caption{
\textbf{Left:} The source image and the mask used for inpainting.
\textbf{Right top:} The generated images based on clean images. 
\textbf{Right down:} The generated images based on adversarial examples. The inpainting district loses some basic structure.
}
\label{inpainting}
\vspace{0.2cm}
\includegraphics[width=0.9\textwidth]{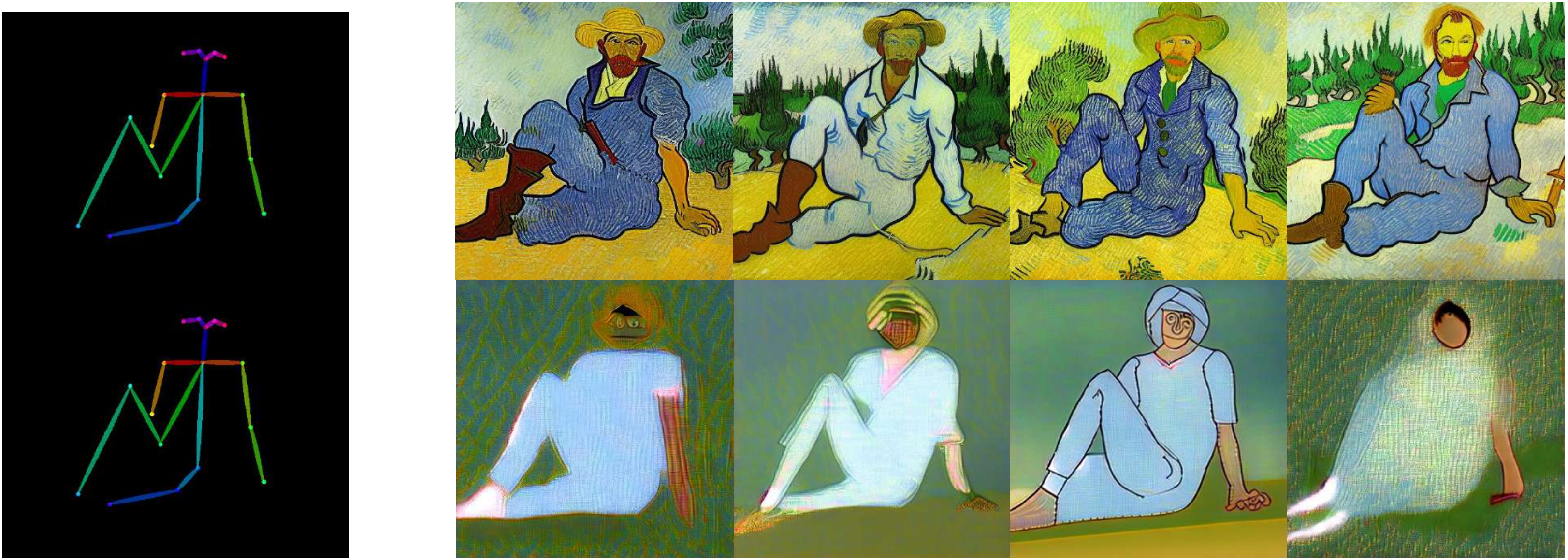}
\caption{
\textbf{Left:} The pose used for generation.
\textbf{Right top:} The generated images based on clean images. 
\textbf{Right down:} The generated images based on adversarial examples. Generated images based on adversarial examples lose the feature of the art style.
}
\label{pose}
\vspace{0.2cm}
\includegraphics[width=0.9\textwidth]{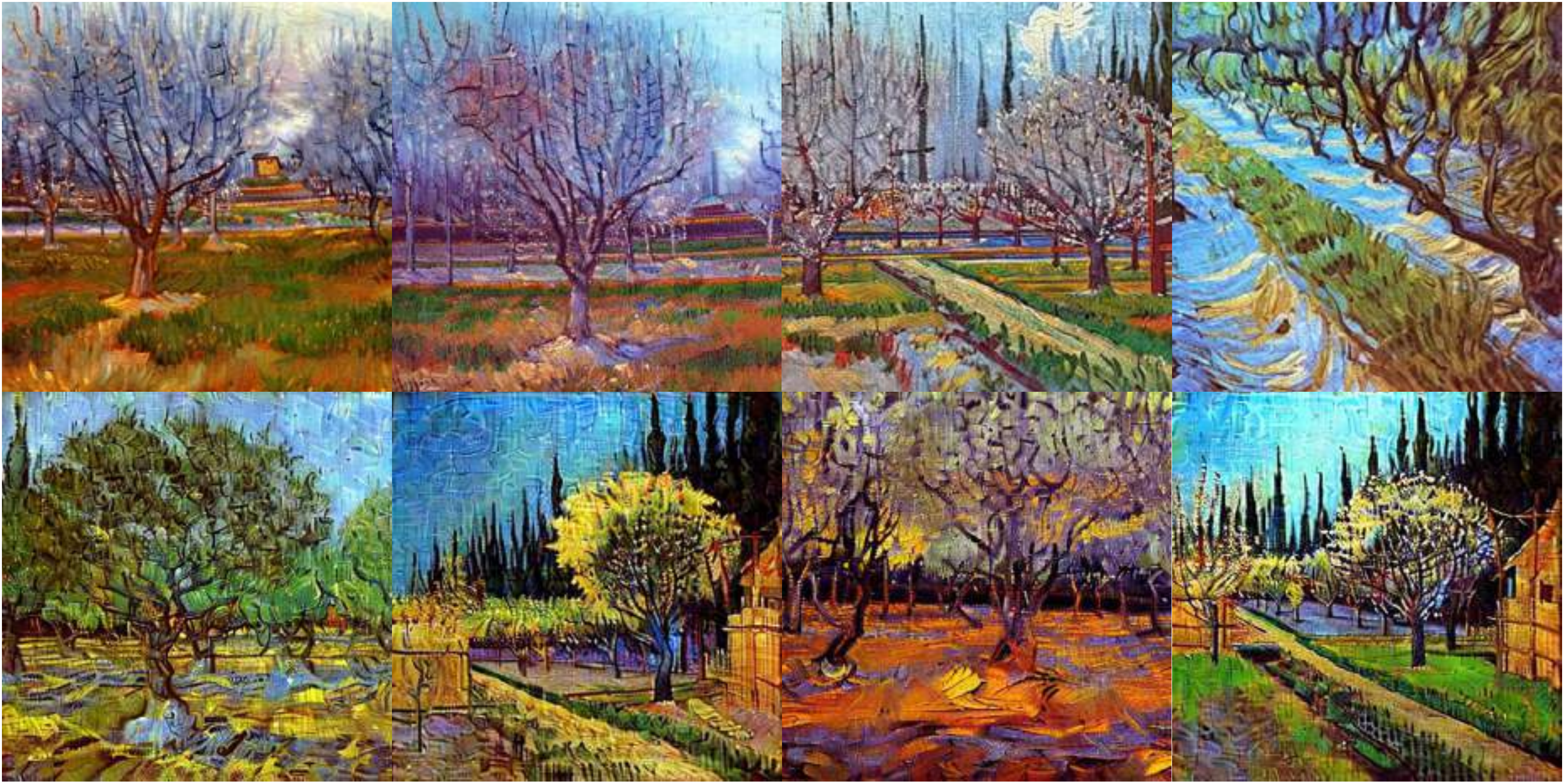}
\caption{
\textbf{Top:} The generated images based on Stable Diffusion trained by DreamBooth using clean images. 
\textbf{Down:} The generated images based on Stable Diffusion trained by DreamBooth using adversarial examples with chaotic textures.
}
\label{dreambooth}
\end{center}
\vspace{-0.2cm}
\end{figure*}

\subsection{AdvDM on other image editing tasks}
In this part, we investigate how AdvDM works on image editing tasks other than textual inversion and image-to-image synthesis. We consider two tasks: inpainting, and pose-guided synthesizing. To show the performance of AdvDM in commercial AI-for-Art applications, we select Stable Diffusion 1.5~\footnote{https://huggingface.co/runwayml/stable-diffusion-v1-5/tree/main} as the backbone to generate and evaluate adversarial examples. Other experimental setups stay consistent with the setup stated in Section 4.

Figure~\ref{inpainting} and Figure~\ref{pose} visualize a case of inpainting and pose-guided synthesis, respectively. The visualization shows that when images are generated based on our adversarial examples, they suffer a bad image quality which makes them not usable. Specifically, the content of the generated image would lose basic structure, show strange artifacts, or be oversimplified.

\subsection{AdvDM on Dreambooth}

In this part, we investigate the performance of AdvDM on Dreambooth, another subject-driven generation method that can be used for art style transfer. We generate adversarial examples with AdvDM on Stable Diffusion 1.5 and evaluate these adversarial examples by conducting style transfer over them with Dreambooth. We use the implementation by the Python library \textit{diffuser}~\footnote{https://github.com/huggingface/diffusers/}. We pick the learning rate as $5\times10^{-6}$ and the number of steps as $4000$. Other experimental setups stay consistent with the setup stated in Section 4.

Figure~\ref{dreambooth} shows a comparison case that tried to mimic the art style of Van Gogh with 20 paintings, the same as the setup stated in Section 4.2. We conduct the mimicry on two groups of images: one group consists of clean paintings and the other consists of adversarial examples based on these clean paintings. The results show that our adversarial examples add chaotic textures to the generated images and thus make the generated images not usable.

\subsection{AdvDM's transferability on scenario.gg}

\begin{figure*}[htbp]
\begin{center}
\includegraphics[width=0.90\textwidth]{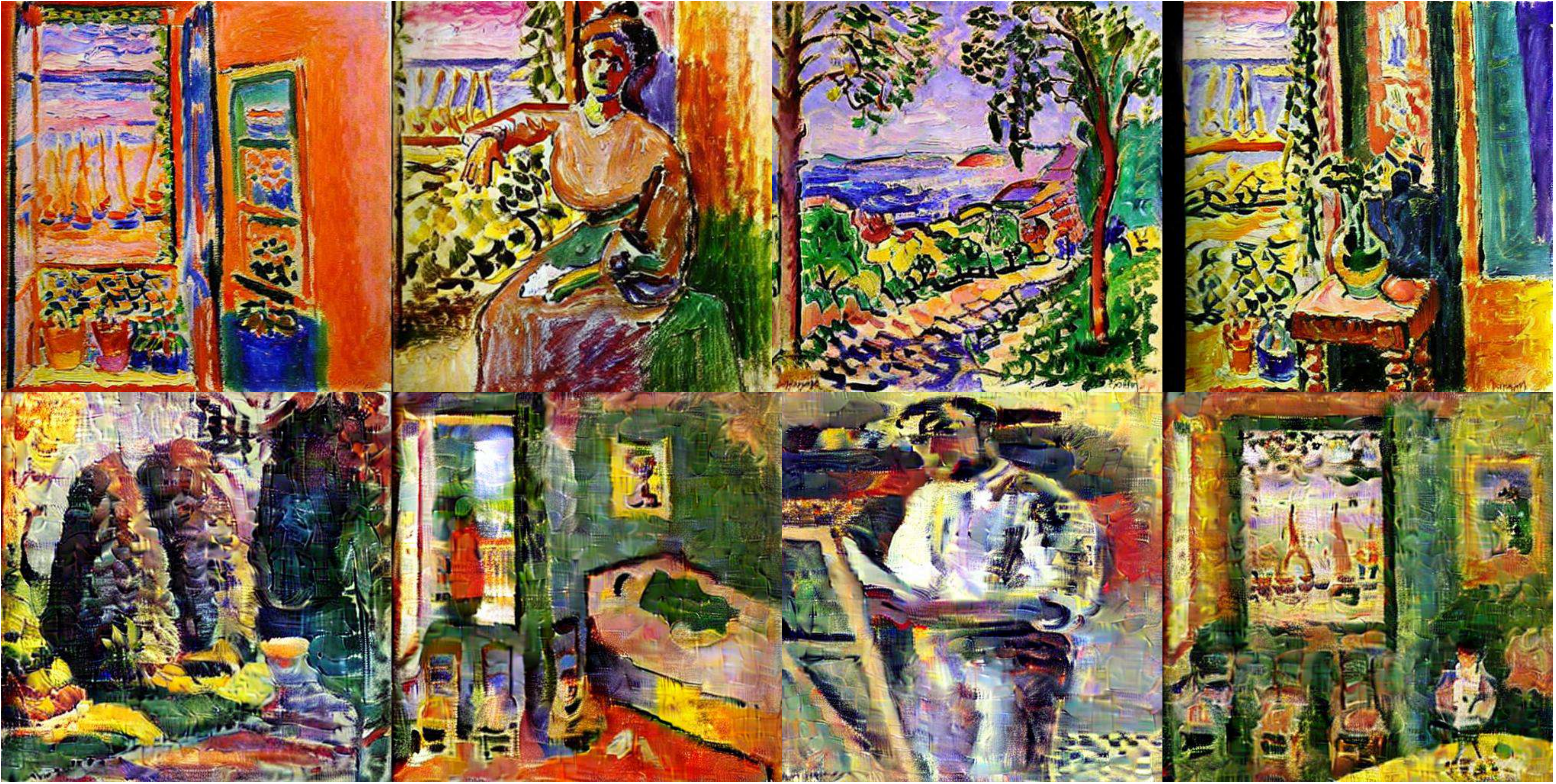}
\caption{
\textbf{The first row:}  Generated images based on clean examples of  Henri Matisse's paintings on commercial AI-for-art websites scenario.gg. \textbf{The second row:} Generated images based on adversarial examples of  Henri Matisse's paintings on scenario.gg. There are chaotic textures on the generated images based on adversarial examples.}
\label{matisse-scenariogg}
\end{center}
\end{figure*}

\begin{figure*}[htbp]

\begin{center}
\includegraphics[width=0.90\textwidth]{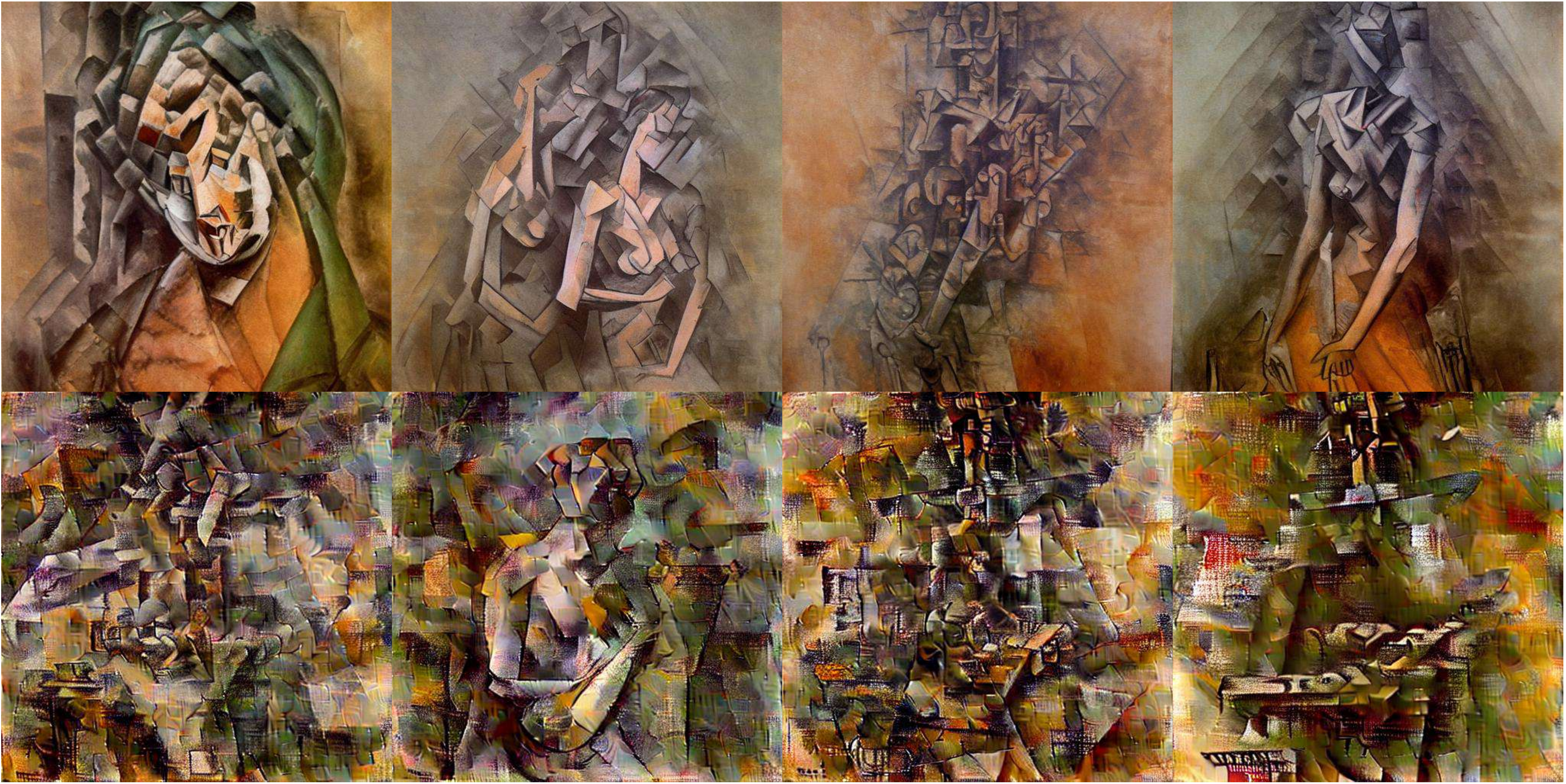}
\caption{
\textbf{The first row:} Generated images based on clean examples of  Pablo Picasso's paintings on commercial AI-for-art websites scenario.gg. \textbf{The second row:} Generated images based on adversarial examples of Pablo Picasso's paintings on scenario.gg. There are chaotic textures on the generated images based on adversarial examples.}
\label{picasso-scenariogg}
\end{center}
\end{figure*}

As a commercial AI-for-Art application that supports art style transfer other than Stable Diffusion, scenario.gg~\footnote{https://app.scenario.gg/} also raises concerns of copyright violation. We conduct experiments to explore whether our adversarial examples can be transferable to scenario.gg. Since scenario.gg is driven by closed-source diffusion models, this experiment aims to investigate the transferability of adversarial examples generated by AdvDM in a black-box adversarial attack setting. For the generation of adversarial examples, the experimental setups stay consistent with the setup stated in Section~\ref{sec:exp}. 

The results are shown in Figure~\ref{matisse-scenariogg} and Figure~\ref{picasso-scenariogg}. Compared to the generated images based on clean images, AdvDM adds chaotic textures to the generated images based on the adversarial examples, though the effect is not as strong as the experiment in the white-box setting. However, the chaotic textures can still make the generated images not usable, which achieves the practical goal of our method.

\subsection{AdvDM against defense: SR and DiffPure}
\label{appendix:diffpure}

\begin{figure*}[htbp]
\begin{center}
\includegraphics[width=0.90\textwidth]{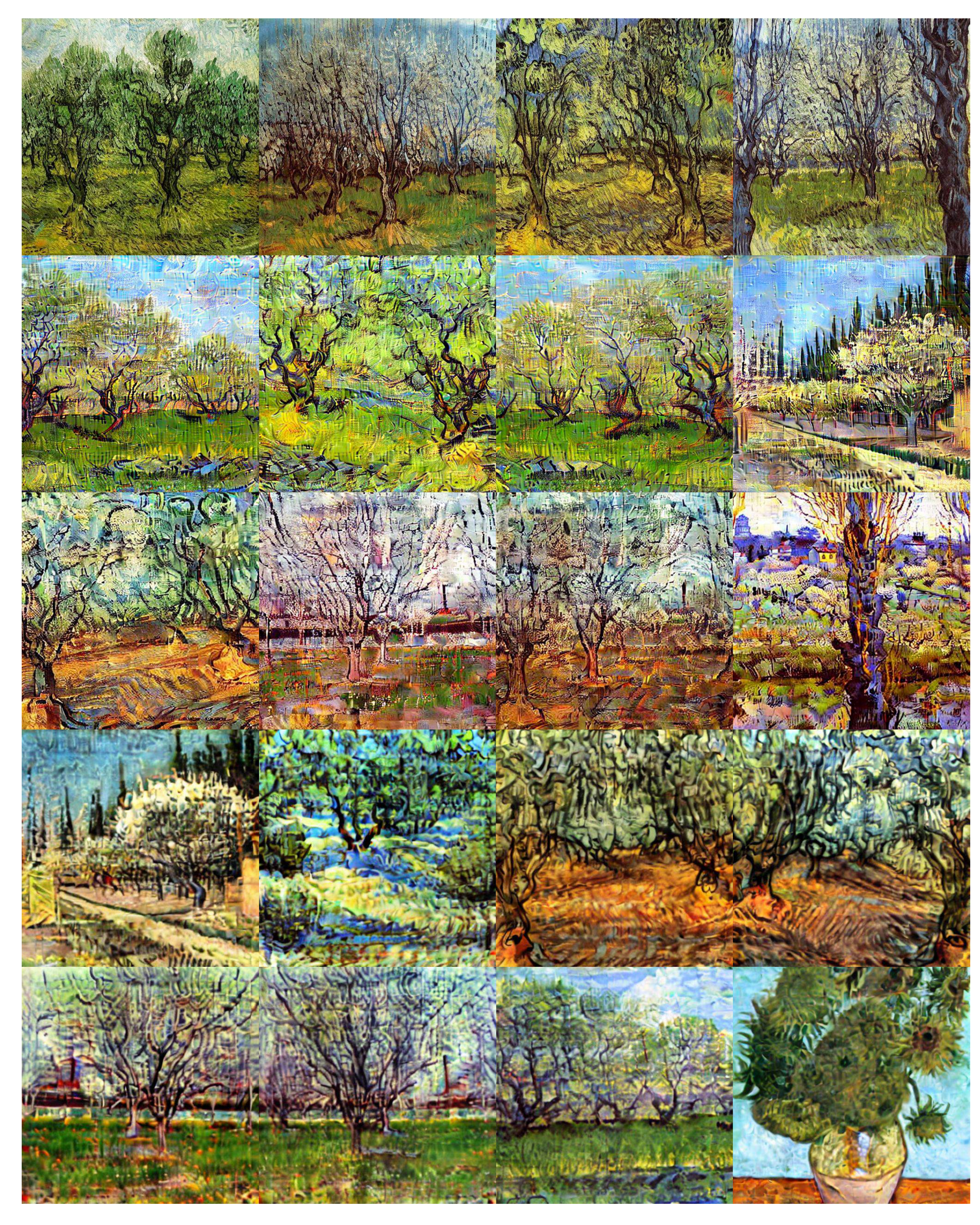}
\caption{Generated images based on clean and adversarial examples of Van Gogh's paintings with SR and DiffPure defenses by scenario.gg.
\textbf{The first row:} clean-example based with no defense. \textbf{The second row:} adversarial-example based with no defense.
\textbf{The third row:} adversarial-example based with SR. \textbf{The fourth row:} clean-example based with SR. \textbf{The fifth row:} adversarial-example based with DiffPure. Both defenses are not able to purify enough adversarial perturbation so that the generated images are still of low quality.}
\label{defense-comparison}
\end{center}
\end{figure*}

One main concern of AdvDM is that its strength may be greatly reduced by the preprocessing-based adversarial defense. In this part, we conduct experiments to illustrate the effectiveness of AdvDM under two state-of-the-art adversarial defenses: SR and DiffPure. To fit the real black box scenario where our method would be applied more, we choose scenario.gg, a commercial AI-for-Art application specific for art style transfer, as the backbone to evaluate the performance of adversarial examples. This can better validate the performance of AdvDM since it is exactly a transfer-learning scenario, as aforementioned in F.3. For SR, we follow the setup stated in Section 4.5. For DiffPure, we utilize the original implementation of DiffPure provided by the author~\footnote{https://github.com/NVlabs/DiffPure}. Note that DiffPure is a model-based noise purification and its effect therefore highly depends on the used model. In the official implementation, the author provides three models, which are trained on Cifar-10, ImageNet, and CelebA-HQ with the image resolution of $32 \times 32$, $224 \times 224$, and $224 \times 224$, respectively. In this experiment, we choose the model trained on ImageNet, for it has a high resolution and the content of the dataset is relatively similar to the content of paintings used in our experiments. All the setups stay as the default setting of the official implementation of DiffPure.

Figure~\ref{defense-comparison} visualizes the results. Both SR and DiffPure are not able to prevent our adversarial examples from adding chaotic textures to the generated images. Specifically, the generated images based on clean examples with DiffPure are also of low quality. This is because the resolution of output images in DiffPure is limited and output images suffer from a reduction in image quality during the process of noise purification.

\end{document}